\documentclass[twoside]{article}

\usepackage[accepted]{aistats2022}

\usepackage[round]{natbib}

\usepackage[utf8]{inputenc} %
\usepackage[T1]{fontenc}    %
\usepackage[hidelinks]{hyperref}       %
\usepackage{url}            %
\usepackage{booktabs}       %
\usepackage{amsmath, amsfonts}       %
\usepackage{nicefrac}       %
\usepackage{microtype}      %
\usepackage{xcolor}         %
\usepackage{braket}

\usepackage{amsmath,amssymb, amsthm}
\usepackage{algorithm}
\usepackage{algpseudocode}
\algtext*{EndFunction}%
\algtext*{EndFor}%
\algtext*{EndIf}%
\usepackage{subfigure}
\usepackage{mathtools}
\usepackage{bm}
\usepackage{soul}

\usepackage{tikz}
\usetikzlibrary{arrows, decorations.pathreplacing}
\usetikzlibrary{patterns}
\usepackage{url}            %
\usepackage{booktabs}       %
\usepackage{nicefrac}       %
\usepackage{braket}
\usepackage{paralist}

\usepackage[shortlabels]{enumitem}
\usepackage[normalem]{ulem}

\newtheorem{prop}{Proposition}
\newtheorem{thm}{Theorem}

\newtheorem{remark}{Remark}

\newcommand{\argmax}{\mathop{\mathrm{argmax}}\limits}
\newcommand{\expect}[2]{{\ensuremath{\mathbb{E}_{#1}\left[{#2}\right]}}}

\newcommand{\xfull}{\ensuremath{\mathbb{X}}}
\newcommand{\xtr}{\ensuremath{\mathbb{X}_\text{tr}}}
\newcommand{\xte}{\ensuremath{\mathbb{X}_\text{te}}}
\newcommand{\yfull}{\ensuremath{\mathbb{Y}}}
\newcommand{\ytr}{\ensuremath{\mathbb{Y}_\text{tr}}}
\newcommand{\yte}{\ensuremath{\mathbb{Y}_\text{te}}}
\newcommand{\zfull}{\ensuremath{\mathbb{Z}}}
\newcommand{\ztr}{\ensuremath{\mathbb{Z}_\text{tr}}}
\newcommand{\zte}{\ensuremath{\mathbb{Z}_\text{te}}}

\newcommand{\ntr}{\ensuremath{n_\text{tr}}}
\newcommand{\nte}{\ensuremath{n_\text{te}}}
\newcommand{\mtr}{\ensuremath{m_\text{tr}}}
\newcommand{\mte}{\ensuremath{m_\text{te}}}

\newcommand{\chol}{\ensuremath{\mathrm{chol}}}
\newcommand{\eye}{\ensuremath{\bm{I}}}

\algnewcommand{\IIf}[1]{\State\algorithmicif\ #1\ \algorithmicthen}
\algnewcommand{\EndIIf}{\unskip\ \algorithmicend\ \algorithmicif}
\algnewcommand{\Ielse}[1]{\unskip\ \algorithmicelse\ #1}

\begin{document}

\twocolumn[

\aistatstitle{A Witness Two-Sample Test}

\aistatsauthor{Jonas M.~Kübler \And Wittawat Jitkrittum \And Bernhard Sch\"olkopf \And Krikamol Muandet}

\aistatsaddress{ Max Planck Institute \\for Intelligent Systems\\Tübingen, Germany
\And Google Research \And  Max Planck Institute \\for Intelligent Systems\\Tübingen, Germany \And  Max Planck Institute \\for Intelligent Systems\\Tübingen, Germany} ]

\begin{abstract}
The Maximum Mean Discrepancy (MMD) has been the state-of-the-art nonparametric test for tackling the two-sample problem. Its statistic is given by the difference in expectations of the witness function, a real-valued function defined as the mean of kernel evaluations on a set of basis points. Typically the kernel is optimized on a training set, and hypothesis testing is performed on a separate test set to avoid overfitting (i.e., control type-I error).  That is, the test set is used to simultaneously estimate the expectations and define the basis points, while the training set only serves to select the kernel and is discarded.  In this work, 
we propose to use the training set to also define the weights and the basis points for better data efficiency.  We show that 1) the new test is consistent and has a well-controlled type-I error; 2) the optimal witness function is given by a precision-weighted mean in the reproducing kernel Hilbert space associated with the kernel;
 and 3) the test power of the proposed test is comparable or exceeds that of the MMD and other modern tests, as verified empirically on challenging synthetic and real problems (e.g., Higgs data).
\end{abstract}

\section{INTRODUCTION}
We tackle the classic \emph{two-sample problem}: given two samples, do they differ significantly enough that we can conclude they originate from two different distributions? This is a common task in many life sciences such as bioinformatics and cancer diagnosis \citep{borgwardt2006bio}. To decide this, one can perform a \emph{two-sample test}, whose goal is to reject the \emph{null hypothesis} "the probability distributions are the same" in favor of the \emph{alternative hypothesis} "the probability distributions are not the same" based on data \citep{Lehmann05:Testing}.
To quantitatively assess this, one defines a \emph{test statistic} and estimates its value on the observed samples. If we know (or are able to simulate) the distribution of this test statistic under the null, we can reject the null if the observed value is significantly larger than what we would expect if the null was true.
Traditional hypothesis tests have test statistics that are defined a priori. A simple example are $t$- or $z$-tests, which only test whether the empirical means of both samples differ significantly \citep{Lehmann05:Testing}. However, such a simple approach is not sufficient to detect differences of distributions with the same mean but, for example, different variance, skewness, or kurtosis.

To detect any differences between two distributions we focus on two categories of tests closely tied to machine learning, but note that various other methods exist \citep{Friedman1979, Chen2017}.
The former first transforms data into a high-dimensional feature space based on a pre-defined feature map, e.g., kernel function. The test statistics can then be defined in terms of the embeddings of the two distributions in the feature space \citep{MouBacHar2008, gretton2012kernel}. The second approach instead learns to distinguish the two distributions by training a classifier, e.g., via a deep neural network. Based on the learned model, the test statistics is then computed on an independent set of samples, e.g., through data splitting \citep{Friedman03:Tests, kim2016classification,LopOqu2017, CheClo2019}.

The popular kernel two-sample test based on the \emph{Maximum Mean Discrepancy} (MMD) in principle does not require data splitting and is completely determined a priori by a positive definite kernel function \citep{gretton2012kernel}.
However, recent research has shown that optimizing the kernel function on a held-out dataset increases the power of the MMD-based tests \citep{Gretton2012optimal, sutherland2016generative,  liu2020learning,KirchlerKKL20}. Thus most modern MMD-based tests are used as two-stage procedures with data splitting, although it is in principle possible to use the entire dataset for kernel selection and testing \cite{pmlr-v23-fromont12, Fromont2013AoS, kbler2020learning}.\footnote{\citet{schrab2021mmd} recently proposed an aggregated MMD two-sample test working without data splitting.}

To obtain maximally significant results in the testing phase, we advocate that in a ``two-stage'' two-sample test, it is more appropriate to learn a test statistic that is as problem-specific as possible.
For the MMD tests, this means that we advocate to 
learn a one-dimensional witness function and not a kernel.
To formalize this, we propose a general two-stage witness two-sample test (WiTS test).
The introduced WiTS test has the following properties:
\begin{itemize}[noitemsep, topsep=0pt, leftmargin=*]
    \item The test statistic is the difference in means of a one-dimensional function called the \emph{witness function} and is thus asymptotically normal under \emph{both} the null and alternative hypotheses. This allows for a simple theoretical treatment (cf. Theorem \ref{thm:asympt_1d} and Proposition \ref{prop:consistency}).
    \item Compared to \citet{sutherland2016generative} and \citet{liu2020learning}, the WiTS test has a simpler test power criterion as a training objective and test thresholds 
    can be simulated more efficiently (cf.~Section \ref{sec:witness_function_based_tests} \& Eq.~\eqref{eq:SNR}).
    \item The WiTS tests empirically outperform the benchmark tests of \citet{liu2020learning} and classification-based tests on challenging synthetic and real problems, e.g., Higgs data (cf. Figure \ref{fig:benchmark}).
\end{itemize}

The rest of the paper is organized as follows.
Section \ref{sec:Motivation} reviews MMD based two-sample tests with a focus on the witness function and discusses our motivation.
We then present the general WiTS test framework in Section \ref{sec:witness_function_based_tests}, followed by a specific example in Section \ref{sec:kfda_witness}.
Next, we discuss related work in detail in Section \ref{sec:related_work}.
Finally, Section \ref{sec:experiments} provides the empirical results comparing the proposed WiTS tests to existing ones on several benchmark datasets. The code to reproduce the experiments is published under \url{https://github.com/jmkuebler/wits-test}.

\section{BACKGROUND AND MOTIVATION}\label{sec:Motivation}
\paragraph{Notation and definitions.}
Let $X, Y$ be random variables with probability distributions $P$ and $Q$ on $\mathcal{X}\subseteq \mathbb{R}^d$, respectively.
In this work, we aim to test the null hypothesis $H_0: P=Q$ against the alternative $H_1: P\neq Q$ based on samples $\xfull = \{x_1,\dots, x_n\}$ and $\yfull=\{y_1, \dots, y_m\}$ drawn i.i.d.~from $P$ and $Q$, respectively.
Rejecting $H_0$ although it is true creates a type-I error, whereas
not rejecting the null when it is false creates a type-II error.
Desirable testing procedures should minimize the type-II error rate, while controlling the type-I error rate at a significance level $\alpha$ (or below). When we consider data splitting, we use $\xtr, \xte$ and $\ytr, \yte$ to denote the disjoint training and test sets with $n=\ntr +\nte$, $m=\mtr + \mte$. We define the shorthands $[n] := \{1 ,\ldots, n\}$, $\zfull = \{\xfull, \yfull\}$,  $\ztr = \{\xtr, \ytr\}$ and $\zte = \{\xte, \yte\}$.

Although most of our analysis applies to more general function spaces, we will
consider a reproducing kernel Hilbert space (RKHS) $\mathcal{H}$ with positive definite kernel $k:
\mathcal{X}\times \mathcal{X} \to \mathbb{R}$ \citep{Scholkopf01:LKS}.
By the Riesz representation theorem, we have that $f(x) = \braket{f, k(x,\cdot)}$ for all $x\in\mathcal{X}$ and $f\in \mathcal{H}$. We assume that $$\textbf{(A1): }\expect{}{k(X,X)} < \infty, \,\expect{}{k(Y,Y)} < \infty$$ holds.
\textbf{(A1)} ensures the kernel mean embeddings  of $P$ and $Q$ exist, i.e., $\mu_P = \expect{}{k(X, \cdot)}, \mu_Q=\expect{}{k(Y,\cdot)}$,  and that we can write $\expect{}{f(X)} = \braket{f, \mu_P}$ for all $f\in \mathcal{H}$ \citep{Muandet2017}.
For a sample $\xfull$, we define the empirical mean embedding as $\mu_\xfull = \frac{1}{|\xfull|}\sum_{x\in\xfull} k(x, \cdot)$.

\paragraph{MMD and witness function.}A popular class of two-sample tests are based on the \emph{Maximum Mean Discrepancy} (MMD) \citep{gretton2012kernel}.
 The MMD of two distributions with respect to the unit ball of $\mathcal{H}$ is defined as \citep[Eq.~(1)]{gretton2012kernel}: $
    \text{MMD} = \sup_{f\in \mathcal{H}, \|f\|\leq 1} \left\{\expect{}{f(X)} -  \expect{}{f(Y)} \right\}.$
The function that \emph{witnesses} the MMD is $\text{argmax}_{f\in \mathcal{H}, \|f\|\leq 1} \{\expect{}{f(X)} -  \expect{}{f(Y)}\} = (\mu_P - \mu_Q) / \|\mu_P - \mu_Q\|$ \citep[Sec.~2.3]{gretton2012kernel}. We define its unnormalized version as $h_k^{P,Q} = \mu_P - \mu_Q$ and obtain
\begin{align}\label{eq:square_MMD}
\begin{aligned}
        \text{MMD}^2 &= \braket{\mu_P-\mu_Q, \mu_P-\mu_Q} = \braket{\mu_P-\mu_Q, h_k^{P,Q}} \\
    &= \expect{}{h_k^{P,Q}(X)} -  \expect{}{h_k^{P,Q}(Y)}.
\end{aligned}
\end{align}
With a \emph{characteristic} kernel \citep{Sriperumbudur2010}, $\mu_P = \mu_Q$ if and only if $P=Q$. Hence, the squared MMD \eqref{eq:square_MMD} can be used to test the hypothesis $H_0: P=Q$ against $H_1: P \neq Q$.

\paragraph{MMD-BOOT test statistics.}
 We can estimate the squared MMD \eqref{eq:square_MMD} by replacing the witness $h_k^{P,Q}$ and the expectations in \eqref{eq:square_MMD} with their empirical counterparts $h_k^{\zfull} = \mu_\xfull - \mu_\yfull$ and
\begin{align}\label{eq:V-statistic}\begin{aligned}
    &\widehat{\text{MMD}}_{\tiny{\textsc{boot}}}^2(\zfull| k) = \frac{1}{n}\sum_{x\in \xfull} h_k^{\zfull}(x) - \frac{1}{m}\sum_{y\in \yfull} h_k^{\zfull}(y) \\
    &= \Braket{ \frac{1}{n} \sum_{x\in \xfull} k(x,\cdot) - \frac{1}{m} \sum_{y\in \yfull} k(y,\cdot), h_k^{\zfull}(\cdot)} \\
    &= \frac{1}{n^2} \sum_{x,x'\in \xfull} k(x,x') + \frac{1}{m^2} \sum_{y,y'\in \yfull} k(y,y') \\
    & \quad - \frac{2}{nm} \sum_{x\in \xfull, y\in\yfull} k(x,y).
\end{aligned}
\end{align}
The latter expression is a sum of $V$-statistics and up to the biased terms where $x=x'$ or $y=y'$ equals the unbiased $U$-statistic which is the standard MMD estimate \citep{gretton2012kernel}.
The witness itself depends on the same data $\zfull$ used to evaluate the test statistic \eqref{eq:V-statistic}
and compute the test threshold, the null distribution has to be simulated via permutation of the samples (or bootstrapping) \citep{gretton2012kernel}.
Thus, we refer to this approach as \textsc{mmd-boot}.\footnote{Our naming convention should emphasizes that the asymptotic distribution cannot be evaluated in closed-form and hence we necessarily need to simulate it. Note, however, that in practice often permutations are used \citep{sutherland2016generative} and it is not necessary to completely simulate the distribution from scratch.}

\paragraph{OPT-MMD-BOOT test statistics.}
A drawback of \textsc{mmd-boot} is that the kernel $k$ has to be chosen a priori before observing the data.
Kernel choice, however, critically affects the performance of MMD based two-sample tests \citep{Gretton2012optimal, sutherland2016generative, liu2020learning, kbler2020learning, JitSzaChwGre2016}.
It is thus common to split the data into $\zfull = (\ztr,\zte)$ and optimize the kernel only on the held-out set $\ztr$.
For the moment, without specifying how the kernel is optimized, we denote the resulting optimized kernel as $k_{\text{tr}}$ with a subscript $\mathrm{tr}$ to indicate that it depends on the training data.
After optimizing the kernel, the standard \textsc{mmd-boot} test is conducted on $\zte$ with the optimized kernel $k_\text{tr}$ \citep{sutherland2016generative, liu2020learning}.
Hence, the empirical expectations and witness function in \eqref{eq:V-statistic} are still dependent on the same data $\zte$, and  the null distribution still has to be bootstrapped, for the same reason as in the case of \textsc{mmd-boot}. We will refer to this approach as \textsc{opt-mmd-boot} with the test statistic
\begin{align}\label{eq:opt-mmd-boot}
\begin{aligned}
    &\widehat{\text{MMD}}_{\tiny{\textsc{opt-boot}}}^2(\zte| k_\text{tr}) \\
    &= \frac{1}{\nte}\sum_{x\in \xte} h_{k_{\text{tr}}}^{\zte}(x) - \frac{1}{\mte}\sum_{y\in \yte} h_{k_{\text{tr}}}^{\zte}(y).
\end{aligned}
\end{align}

\paragraph{Our Motivation.}

This is the starting point of our investigations: Although the kernel is optimized, it is still a multidimensional representation of the data. While this makes the test statistic applicable to other problems  \citep{liu2020learning, KirchlerKKL20}, features that contain little information about the differences of $P$ and $Q$ will mainly add noise to the test statistic. Generally, the noisier the test statistic, the harder it is to obtain  significant test results.
Motivated by this drawback, we propose to formulate a test statistic that is more specific to the observed difference in $\ztr$. Being more specific to the training data (that is all we know about $P$ and $Q$), comes at the risk of overfitting, which we mitigate via regularization and model selection (cf.~Section \ref{sec:witness_function_based_tests}). Specifically for MMD, after the kernel is optimized, \emph{we define the witness directly on the training data} by replacing $h_{k_{\text{tr}}}^{\zte}$ with $h_{k_{\text{tr}}}^{\ztr} = \frac{1}{\ntr} \sum_{x\in\xtr} k_\text{tr}(x, \cdot) - \frac{1}{\mtr} \sum_{y\in\ytr}k_\text{tr}(y, \cdot)$.
We call this \textsc{opt-mmd-witness}:
\begin{align}\label{eq:opt-mmd-witness}
\begin{aligned}
    &\widehat{\text{MMD}}_{\tiny{\textsc{opt-witness}}}^2\left(\zte|h_{k_{\text{tr}}}^{\ztr}\right) \\
    &= \frac{1}{\nte}\sum_{x\in \xte} h_{k_{\text{tr}}}^{\ztr}(x) - \frac{1}{\mte}\sum_{y\in \yte} h_{k_{\text{tr}}}^{\ztr}(y).
\end{aligned}
\end{align}
This test statistic comes with numerous advantages.
Firstly, the expectations (defined via $\zte$) are now independent of the witness function (defined via $\ztr$). Thus, the test statistic is asymptotically normal.
Secondly, as we will see in the following sections, \eqref{eq:opt-mmd-witness} allows us to compute asymptotic test thresholds in closed form and allows for a simpler derivation of a test power criterion than in the case of \textsc{opt-mmd-boot} \citep{sutherland2016generative, liu2020learning}.
Lastly, our empirical results suggest that \textsc{opt-mmd-witness} outperforms \textsc{opt-mmd-boot} on datasets considered in \citet{liu2020learning}.

\section{WITNESS TWO-SAMPLE TEST (WiTS TEST)}\label{sec:witness_function_based_tests}

Similar to \eqref{eq:opt-mmd-witness}, the WiTS tests we propose are by design two-stage procedures: In \emph{Stage I}, we learn the witness function $h$ with the training data $\ztr$. This ensures that $h$ is independent of  the test data $\zte$, used in \emph{Stage II} to define a test statistic
\begin{align}\label{eq:tau_diff_mean}
    \hat{\tau}(\zte | h) \propto \frac{1}{\nte}\sum_{x\in \xte} h(x) - \frac{1}{\mte}\sum_{y\in \yte} h(y).
\end{align}
We reject the null hypothesis $H_0: P=Q$ if the observed value is larger than a test threshold.
We start presenting Stage II and analyze the test's asymptotic power for a given function $h$.
Then, we will use this test power criterion as the objective when optimizing the witness function in Stage I.

\subsection{Stage II - Testing with the Witness Function}
We start with a basic result on asymptotic normality of empirical means (\citep{serfling1980approximation}, Proof in App.~\ref{app:proof_normality}).
\begin{thm}[Asymptotic normality of WiTS test]\label{thm:asympt_1d}
For a witness function $h : \mathcal{X} \to \mathbb{R}$, let $\sigma_P^2 := \text{Var}[h(X)]$ and $\sigma_Q^2 := \text{Var}[h(Y)]$ such that $0 < \sigma^2_P,\sigma^2_Q < \infty$.
Let $\{X_i\}_{i\in[n]} \overset{\text{i.i.d.}}{\sim} P$, $\{Y_j\}_{j\in[m]}
\overset{\text{i.i.d.}}{\sim} Q$, and $c:= \frac{n}{n+m} \in (0,1)$
as $n+m \to \infty$.
Denote by $\bar{h}_P := \expect{}{h(X)}$ and
$\bar{h}_Q :=\expect{}{h(Y)}$. We define the empirical means $\hat{h}^n_P := \frac{1}{n} \sum_{i\in [n]} h(X_i)$, $\hat{h}^m_Q := \frac{1}{m} \sum_{i\in [m]} h(Y_i)$ and denote the sample variance as
$\hat\sigma^2_c(h) :=\hat\sigma^2_P/ c + \hat\sigma^2_Q/(1-c)$. Then
\begin{align*}
    \frac{\sqrt{n+m}}{\hat\sigma_c(h)} \left[\left( \hat{h}^n_P- \bar{h}_P \right) - \left(  \hat{h}^m_Q - \bar{h}_Q\right) \right]
    \overset{d}{\to} \mathcal{N} \left(0, 1\right).
\end{align*}
\end{thm}
For any \textit{fixed} $h$ and for sufficiently large sample sizes, we can thus work with the asymptotic distribution of test statistics of the form $\tau(\cdot |h)$ in  Eq.~\eqref{eq:tau_diff_mean} to compute test thresholds and derive an asymptotic test-power objective for choosing $h$ based on the training data $\ztr$ in Stage I.
Data splitting ensures that $h$ is independent of $\zte$, which is necessary for Theorem \ref{thm:asympt_1d} to hold.
In the following, to make the comparison between different choices of $h$ easier, we consider the standardized test statistic on the test samples $\zte$
\begin{align*}
\begin{aligned}
    &\tau(\zte | h) =\sqrt{\nte + \mte}\frac{\frac{1}{\nte} \sum\limits_{x\in \xte}h(x) - \frac{1}{\mte} \sum\limits_{y\in \yte} h(y)}{\hat\sigma_c(h)},
\end{aligned}
\end{align*}
where $c = \frac{\nte}{\nte + \mte}$ and $\hat\sigma_c(h)$ is the empirical estimate of the pooled variance as in Theorem \ref{thm:asympt_1d} based on $\zte$.
To control the type-I error at a significance level $\alpha$, we need to find a \emph{test threshold} $t_\alpha$ such that $P(\tau(\zte | h) > t_\alpha | H_0) \leq \alpha$.
By Theorem \ref{thm:asympt_1d}, we can define the threshold to be the $(1-\alpha)$ quantile of the asymptotic null distribution.
Under the null hypothesis we have $\bar{h}_P = \bar{h}_Q$ and obtain $t_\alpha = \Phi^{-1}(1-\alpha)$ where $\Phi^{-1}$ denotes the inverse CDF of the standard normal.

Note that we only consider a "one-sided" test, since we
choose $h$ in stage I with the appropriate sign, i.e., such that it has larger
expectation under $\xtr$ than under $\ytr$. A "two-sided" test ignores this and
may lead to a reduction in test power.

We reject the null hypothesis $H_0:P=Q$ if $\tau(\zte | h) > t_\alpha$.
As an advantage of the asymptotic normality under the alternative and the closed
form of the threshold of our test, we can write the asymptotic
type-II error rate in closed form, similar as in \citet[Sec.~3]{Gretton2012optimal}:
\begin{align}\label{eq:power}
\begin{aligned}
    &P(\tau(\zte | h) < t_\alpha) \\
    &\approx \Phi\left( \Phi^{-1}(1-\alpha) - \sqrt{\nte+\mte} \, \frac{\bar{h}_P - \bar{h}_Q}{\sigma_c(h)}\right).
\end{aligned}
\end{align}

An important consideration in designing a hypothesis test is test consistency.
A hypothesis test is called consistent, if for a fixed alternative hypothesis, its test power converges to one as sample size goes to infinity.
With \eqref{eq:power}, we can characterize for which functions $h$ the statistic $\tau_h$ leads to a consistent test.
\begin{prop}[Consistency of WiTS test]\label{prop:consistency}
    Assume $0< \sigma_c(h)<\infty$, where $\sigma_c(h)$ is defined in Theorem \ref{thm:asympt_1d}.
    A WiTS test based on
    $h$ is consistent against a fixed alternative hypothesis $P\neq Q$ if and only if $\bar{h}_P > \bar{h}_Q$.
\end{prop}

Proposition \ref{prop:consistency}  ensures that,
for a given alternative hypothesis, our proposed test will eventually (in the
limit of the sample size) reject the null hypothesis $H_0$ when it is false.
Associated with this notion is the \emph{test power}, the probability that
the test rejects $H_0$ when it is false; this quantity is equivalent to $1-$ type-II error.
Defining the \emph{signal-to-noise} ratio $\text{SNR}(h)=\frac{\bar{h}_P - \bar{h}_Q}{\sigma_c(h)}$, it follows from \eqref{eq:power} that the asymptotic test power of our test
is
\begin{align}\label{eq:SNR}
    \beta_h \approx 1 - \Phi\left( \Phi^{-1}(1-\alpha) - \sqrt{\nte+\mte} \, \text{SNR}(h)\right).
\end{align}

Since $\Phi$ increases  monotonically, the test power grows monotonically with the \emph{signal-to-noise} ratio (SNR).

\subsection{Stage I - Finding an Optimal Witness}
We now propose an objective to find an optimal witness function.
Based on our test power consideration, we argue that in the first stage one
should find a witness by maximizing a, possibly regularized, empirical estimate
of the SNR in \eqref{eq:SNR}.
Let $\mathcal{F}$ be a function class containing candidates for the witness.
We propose using the witness $\hat{h}_\lambda$ defined as
\begin{align}
  \label{eq:general_objective}
  \begin{aligned}
    &\hat{h}_\lambda = \argmax_{f\in \mathcal{F}} \frac{\bar{f}_{\xtr} - \bar{f}_{\ytr}}{\sigma^{\ztr}_{c, \lambda}(f)},\\
    &\text{with } \bar{f}_{\xtr} = \frac{1}{\ntr}\sum_{x \in \xtr} f(x), \; \bar{f}_{\ytr} = \frac{1}{\mtr}\sum_{y \in \ytr} f(y),
  \end{aligned}
\end{align}
and $\sigma_{c, \lambda}^{\ztr}(f) = ((\sigma_c^{\ztr}(f)^2 + \lambda \Omega(f))^\frac{1}{2}$, where $\sigma_c^{\ztr}(f)$ corresponds to $\hat{\sigma}_c(h)$ defined in Theorem \ref{thm:asympt_1d} and $\Omega$ is a regularizer.
We remark that the  optimal witness is generally not uniquely defined since the
SNR is invariant to rescaling the function.
Correctly rejecting $H_0$ when it is false is at the core of hypothesis testing.
Our choice of maximizing the SNR in \eqref{eq:SNR} is in line with this principle:
it leads to a test that maximizes the asymptotic test power.
By contrast, while other objectives such as classification
loss\citep{kim2016classification,LopOqu2017}, softmax loss \citep{CheClo2019},
or the MMD statistic itself \citep{gretton2012kernel}, can be used to learn the
witness function, their relationship to the test power may be indirect.

\paragraph{OPT-MMD-Witness.}
A closely related objective to our SNR in \eqref{eq:SNR} was used in
previous work \citep{sutherland2016generative, liu2020learning} to find a good
kernel for a \textsc{opt-mmd-boot} test, see \eqref{eq:opt-mmd-boot}.
For a given kernel $k$, \citet[Eq.(3)]{liu2020learning} derive the training
objective as $J(P,Q|k) = \text{MMD}^2(P,Q|k)/\sigma_{H_1}(P,Q|k)$ where
$\sigma^2_{H_1}(P,Q|k)$ is the asymptotic variance of the MMD estimate under the
alternative hypothesis. In Appendix \ref{proof:u_stat_mmd_snr}, we examine
this quantity in more detail, and show that $J(P,Q|k) =
1/\sqrt{2}\,\text{SNR}(h_k^{P,Q})$. For a given class of kernels and
corresponding (empirical) MMD witnesses, this implies that selecting the optimal
witness according to our SNR criterion leads to the same function as first
optimizing the kernel with the $J$ criterion and defining the witness
afterwards.

\begin{algorithm*}[t]
    \caption{WiTS test with \textsc{kfda-witness}}\label{alg:kfda_witness_test}
    \begin{minipage}{0.45\linewidth}{
	\begin{algorithmic}[1]
        \State \textbf{Input:} $\xfull, \yfull,$ $\alpha$, paramGrid, $r$
        \State $\xtr, \xte, \ytr, \yte \gets$ \Call{RandomSplit}{$\xfull,\yfull,r$}
        \State $\#$ Optionally perform model selection
        \State $k, \lambda \gets$  \Call{GridSearchCV}{paramGrid, $\ztr$}
        \State $\#$ Stage I - Optimize Witness
        \State $h \gets$ \Call{kfdaWitness}{$\ztr, k, \lambda$} \Comment{App. Alg.\ref{Alg:KFDA_Implementation}}
        \State $\#$ Stage II - Test
        \State \textbf{return:} \Call{witnessTest}{$\zte, h, \alpha$}
	\end{algorithmic}
		}
	\end{minipage}~~%
	\begin{minipage}{0.54\linewidth}
		\begin{algorithmic}[1]
			\setcounter{ALG@line}{8}
              \Function{witnessTest}{$\zte, h(\cdot), \alpha$, $B=200$}
              \State $h_{\zte} \gets [h(z) \text{ for $z$ in } \zte ]$
              \State $\tau \gets \Call{mean}{h_{\zte}[:\nte]} - \Call{mean}{h_{\zte}[\nte:]}$
              \State $p \gets 0$ \Comment{simulate $p$-value via permutations}
              \For{$i$ in $[B]$}
              \State $h_{\zte} \gets \Call{Permute}{h_{\zte}}$
              \If{$ \Call{mean}{h_{\zte}[:\nte]} - \Call{mean}{h_{\zte}[\nte:]} \geq \tau$}
              \State $p \gets p + 1/B$
              \EndIf
              \EndFor
              \IIf{$p\leq \alpha$} return: 1 \Ielse return: 0   %
              \EndFunction
		\end{algorithmic}
		\end{minipage}%
\end{algorithm*}

\paragraph{Model Selection and Optimization.}
The choice of function class $\mathcal{F}$ and regularization parameter
$\lambda$ affects the learned witness in \eqref{eq:general_objective}. We
therefore recommend that practitioners use standard tools for model selection
such as cross-validation (CV) for finding suitable ``hyperparameters'' and to
validate that the learned witness actually has a high SNR.
CV ensures that the witness actually learns the differences between $P$ and $Q$ and does not solely overfit the training data.
Model-selection on $\ztr$ is legit since in Stage II we only use $\zte$, which are independent of $\ztr.$ While this is also possible in classifier two-sample tests \citep{LopOqu2017}, in the standard \textsc{mmd-boot} this is not done.

Our objective \eqref{eq:SNR} can be used with a variety of function
classes $\mathcal{F}$. For instance, $\mathcal{F}$ can be  defined based on an
RKHS, or parameterized by a deep neural network. Note that optimization methods
to maximize \eqref{eq:general_objective} are generally function class specific,
and may require an iterative procedure.
However, when $\mathcal{F}$ is an RKHS, we can derive the closed-form solution to
\eqref{eq:general_objective}, as shall be explained in Section
\ref{sec:kfda_witness}. Algorithm \ref{alg:kfda_witness_test} shows the general
procedure for the two-stage WiTS test.

\paragraph{Permutation-based Thresholds.}
For our theoretical analysis we used the asymptotic threshold. However, the witness is also chosen in a data-dependent manner. 
Thus, we generally recommend to simulate the threshold via \textit{permutations} in order to ensure type-I error control at finite sample size. In this case, for simplicity and ease of implementation, we compute the test statistic without normalization and simply take the difference in means. We first compute the value of the witness function on all points in $\zte$ and store it in an array. Then we compute the simplified test statistic by taking the difference in means of $\xte$ and $\yte$ (as computed from the array that stores all the witness evaluations). We then iterate over $B\in \mathbb{N}$ permutation runs to estimate the $p$-value of the computed test statistic. For each run, we permute the array storing the witness evaluations, and then compute the difference in means of the first $\nte$ and the last $\mte$ entries. If this is larger or equal than the test statistic on the original partition, this contributes $1/B$ to the p-value. After all permutations, if the p-value is smaller than $\alpha$, we reject (see Alg.~\ref{alg:kfda_witness_test}). This correctly controls type-I errors, as under the null, the initial partition can be thought of as being itself a random permutation of the data.
Since for this procedure we only need to compute the witness once on each data point the overall cost is $\mathcal{O}((\nte+\mte)B)$. Note that simulating the null for \textsc{mmd-boot} instead has cost $\mathcal{O}((\nte+\mte)^2B)$ \citep[Sec.~5]{liu2020learning}.

\section{KFDA-WITNESS}\label{sec:kfda_witness}
In this section, we consider the function class in
\eqref{eq:general_objective} to be an RKHS, and show that this choice leads
to a closed form solution for the optimal witness.
To start, let $\mathcal{H}$ be an RKHS associated with a positive definite
kernel $k$ (see Section \ref{sec:Motivation}). Additionally to the mean embeddings
$\mu_P, \mu_Q$, we define the (centered) covariance operator $\Sigma_P =
\expect{}{k(X,\cdot)\otimes k(X,\cdot)} - \mu_P \otimes \mu_P$ (analogously for
$Q$) whose existence is ensured by Assumption \textbf{(A1)}
\citep[Sec.~3]{Muandet2017}.
For any function in the RKHS we then have $\expect{}{f(X)} = \braket{\mu_P, f}$ and $\text{Var}[f(X)] = \braket{f, \Sigma_P f}$, and analogously for $Q$.
We define the pooled covariance operator $\Sigma = \frac{\Sigma_P}{c} + \frac{\Sigma_Q}{1-c}$. Then for all $f \in \mathcal{H}$ with non-zero variance we have
\begin{align}\label{eq:population_objective}
    \text{SNR}(f) = \frac{\braket{\mu_P - \mu_Q, f}}{\braket{f, \Sigma f}^\frac{1}{2}},
\end{align}
where $\text{SNR}$ is defined in \eqref{eq:SNR}.
This objective corresponds to Kernel Fisher discriminant analysis (KFDA)'s
learning objective \citep{Mika99}. For singular covariance operator the SNR can diverge, and for infinite-dimensional RKHS, the empirical estimation of the covariance operator is ill-posed. In the following, we therefore consider a regularized ($\lambda >0$) version of \eqref{eq:population_objective} and call its solution \emph{(regularized) KFDA witness}:
\begin{align}\label{eq:reg_witness}
    h_\lambda = \argmax_{\substack{f \in \mathcal{H}}} \frac{\braket{\mu_P - \mu_Q, f}}{\braket{f, (\Sigma + \lambda \mathop{I}) f}^\frac{1}{2}}.
\end{align}
The solution of \eqref{eq:reg_witness} is given by the solution to the generalized eigenvalue problem $(\Sigma + \lambda \mathop{I}) h_\lambda = \gamma   (\mu_P - \mu_Q)$ \citep[Sec.3.2]{Mika_PhD2003}, thus
\begin{align}\label{eq:reg_fisher_witness_explicit}
    h_\lambda = \gamma  (\Sigma + \lambda \mathop{I})^{-1} (\mu_P - \mu_Q),
\end{align}
where $\gamma > 0$ is an arbitrary positive constant we fix to $1$, unless stated otherwise.
We will refer to the test with the witness function $h_\lambda$ as
the \textsc{kfda-witness} test.

Next, we show how we can estimate the KFDA-witness with the training data.

\paragraph{Estimation of the KFDA Witness.}
Let $\mathbb{Z}_\text{tr} = \{x_1,\dots,x_{\ntr}, y_1, \dots, y_{\mtr}\}$ denote the pooled training sample and $K$ denote the kernel matrix such that $K_{ij} = k(z_i,z_j)$ for $i,j \in [\ntr+\mtr]$.
Further, we define $\delta = (\frac{1}{\ntr}, \dots, \frac{1}{\ntr},-\frac{1}{\mtr},\dots, -\frac{1}{\mtr})^\top\in \mathbb{R}^{\ntr+\mtr}$.
For $l \in \{\ntr,\mtr\}$, we define the idempotent centering matrix $P_l =
\mathop{I}_l - l^{-1} \bm{1}_l\bm{1}_l^\top$, where $\mathop{I}_l$ denotes the
identity operator and $\bm{1}_l$ the $l$ dimensional vector with all ones. With
this we define the $(\ntr+\mtr) \times (\ntr+ \mtr) $ matrix
$
    N_c = \begin{pmatrix}
    \frac{1}{c} P_{\ntr} & 0\\ 0 & \frac{1}{1-c}P_{\mtr}
    \end{pmatrix}.$
Using the representer theorem \citep{Scholkopf01:Representer}, we can empirically estimate the KFDA witness (more detail in App.~\ref{app:witness_solution}) as
\begin{align}\label{eq:witness_solution}
    &\hat{h}_\lambda(\cdot) = \sum_{i=1}^{\ntr+\mtr} \hat{\alpha}_i k(z_i, \cdot), \\
    & \hat{\alpha} = \left(\frac{KN_cK}{\ntr +\mtr} + \lambda K\right)^{-1} K\delta.
\end{align}
$\hat{h}_\lambda(\cdot)$ can be viewed as a precision-weighted (inverse covariance) mean of the embeddings of the basis points $\ztr$ in the RKHS.
Since $\mu_{\xtr}, \mu_{\ytr},$ and $\hat{\Sigma}$ are consistent estimates of $\mu_P, \mu_Q$, and $\Sigma$, for fixed regularization, we have $\hat{h}_\lambda \to h_\lambda = (\Sigma + \lambda \mathop{I})^{-1} (\mu_P - \mu_Q)$ (see Appendix \ref{app:convergence}).
For the asymptotic witness $h_\lambda$ we can compute the difference in expectation under $P$ and $Q$ in closed form: $\bar{h}_{\lambda,P} - \bar{h}_{\lambda,Q} = \braket{\mu_P - \mu_Q, h_\lambda} = \braket{\mu_P - \mu_Q, (\Sigma + \lambda \mathop{I})^{-1} (\mu_P - \mu_Q)}$. This difference is positive, and hence by Proposition \ref{prop:consistency} we obtain a consistent WiTS test, if and only if $\mu_P\neq\mu_Q$. We can ensure this for arbitrary $P\neq Q$ by using a \emph{characteristic} kernel \citep{Sriperumbudur2010}, the same condition as for MMD based tests.

\begin{figure*}[t]
    \centering
    \subfigure{\includegraphics[width=.32\linewidth]{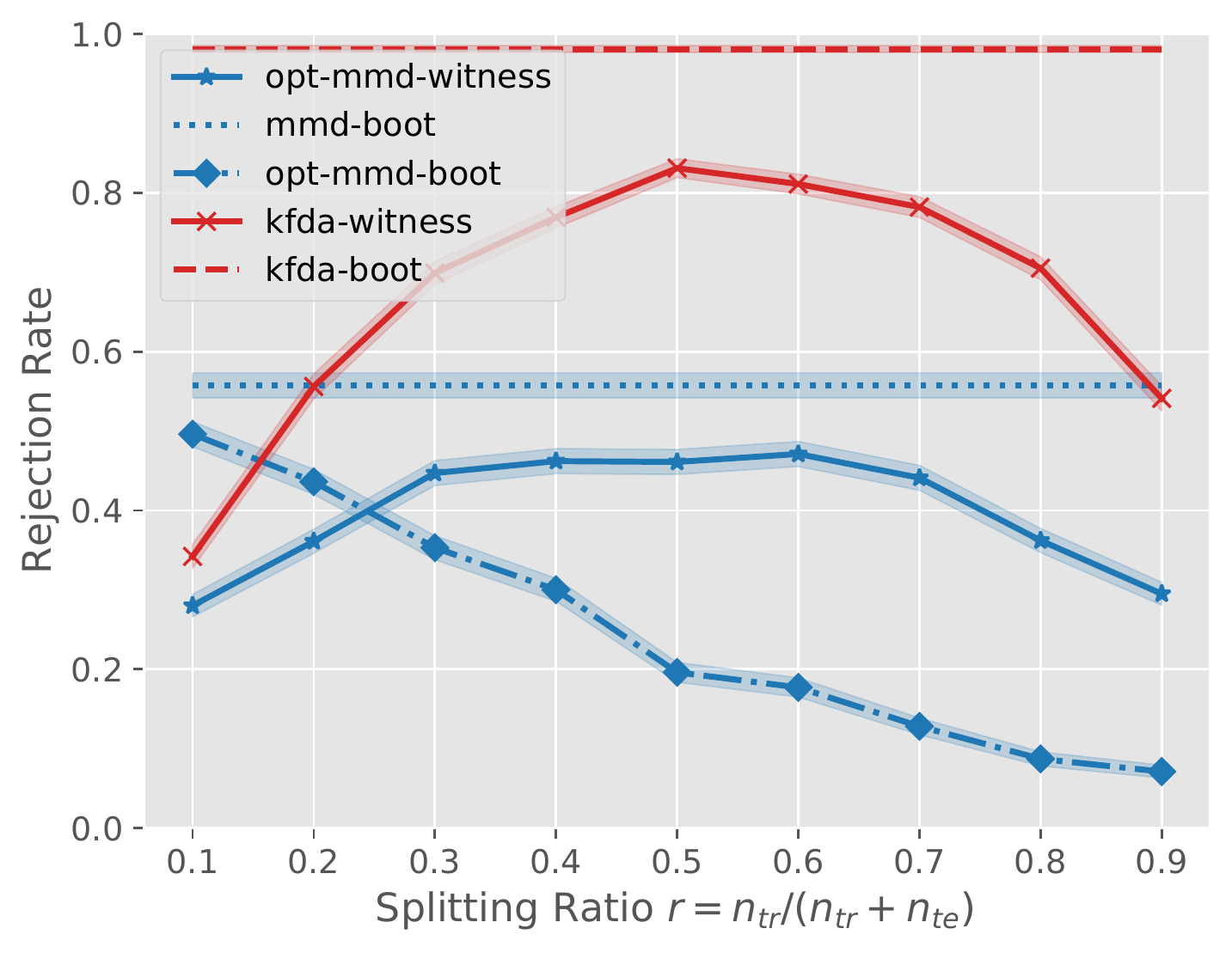}}
    \centering
    \hfill
    \subfigure{\includegraphics[width=.32\linewidth]{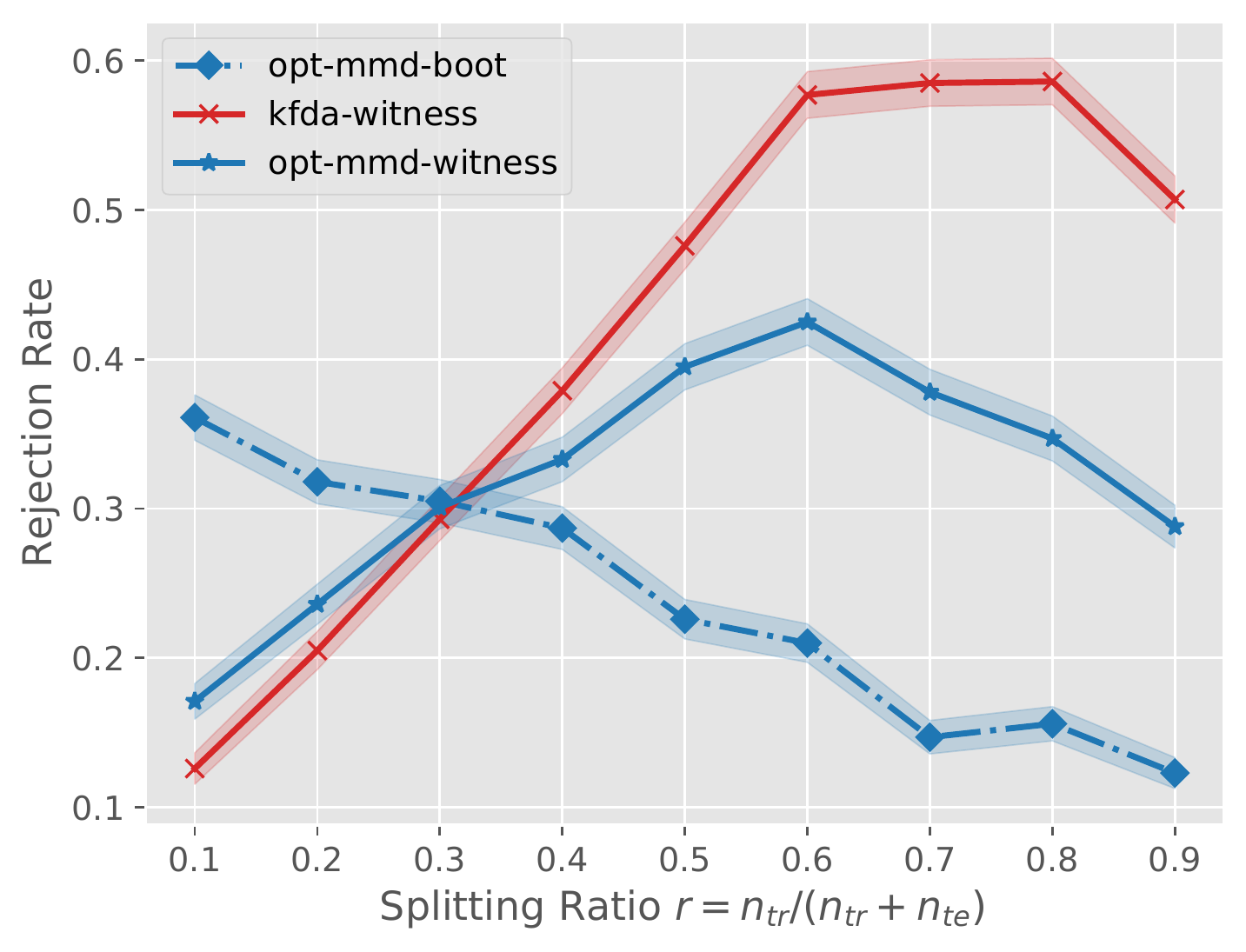}}
    \centering
    \hfill
    \subfigure{\includegraphics[width=.32\linewidth]{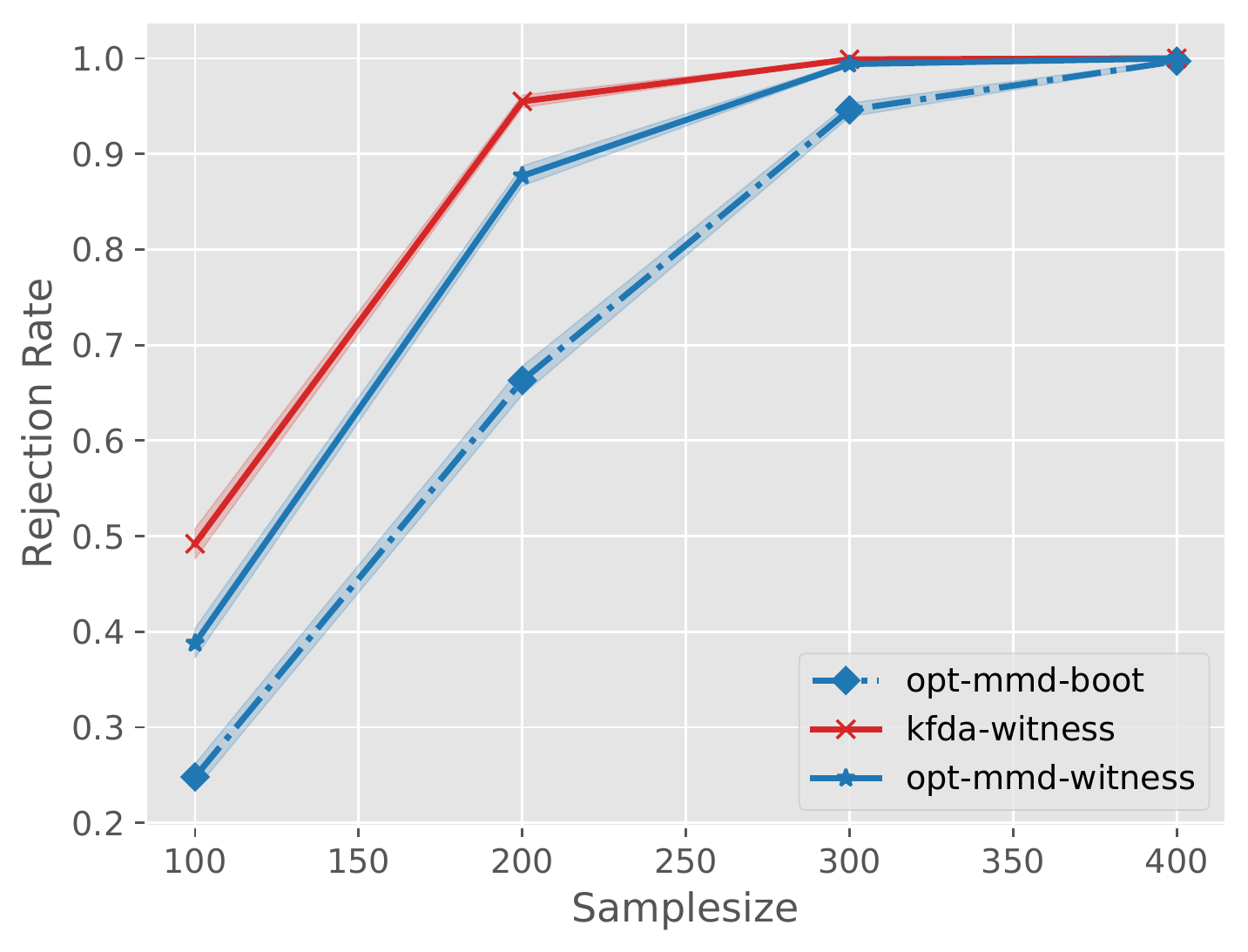}}
    \caption{Instructive experiments on "Blobs" dataset. \textbf{Left:} Fixed kernel and fixed regularization for sample size $n=m=100$. \textbf{Middle:} For multiple candidate kernels ($\mathcal{K}_{10}$) kernel optimization becomes more important and the difference of \textsc{kfda-witness} and \textsc{opt-mmd-witness} becomes smaller. Further, \textsc{opt-mmd-witness} already outperforms \textsc{opt-mmd-boot}. \textbf{Right:} Same kernels as in the middle figure and $r=1/2$. All the tests are consistent, i.e., converge to power equal 1.}
    \label{Fig:instructive}
\end{figure*}

Despite asymptotic consistency, the test power at finite sample size depends on the \emph{splitting ratio} $r\in (0,1)$, i.e., $\ntr = \lceil rn \rceil$ and $\nte = n - \ntr$ and accordingly for the sample from $Q$.
Based on our experimental results, we observe that, for a fixed kernel $k$,
fixed regularization $\lambda >0$ and sufficiently large sample size, the
splitting ratio $r=1/2$ appears to give the highest test power in many cases, compared to other values of $r$. Generally, identifying the optimal
splitting ratio remains an open problem. We observe (middle panel of Fig.~\ref{Fig:instructive})
that if we include model selection in stage I, it is favorable to use more than
half of the data for the first stage, i.e., $r>1/2$. However, since we cannot
quantify how much "\emph{more}" data we should use, we generally recommend using a 50/50 split.

The cost of computing the exact solution $\hat\alpha$ in \eqref{eq:witness_solution} is $\mathcal{O}((\ntr+\mtr)^2)$ in space (storing the kernel matrix) and $\mathcal{O}((\ntr+\mtr)^3)$ time (matrix inversion).
In Appendix \ref{app:KFDA_Falkon}, we adopt recent advances in large-scale kernel machines \citep{Rudi2017, Meanti2020} to obtain approximate solutions with lower time and space complexity and thus scale to large datasets. Using the Nyström approximation \citep{WilliamsS00} to approximate the solution and approximately solving it with conjugate gradient, we obtain a complexity of $\mathcal{O}((\ntr+\mtr )Mt + M^3)$ in time and $\mathcal{O}(M^2)$ in space, where $M$ denotes the number of Nyström centers and $t$ the number of conjugate gradient iterations.
For stage II we then only need $(\nte+\mte) M$ kernel evaluations to compute the test statistic. This makes our approach scalable to large-scale dataset.\footnote{After acceptance of this work, \citet{chatalic2022nystr} proposed a Nystr\"om approximation of the kernel mean embedding to speed up the MMD estimation.}

\paragraph{Connection of \textsc{opt-mmd-witness} and \textsc{kfda-witness}.}
To emphasize the relationship between optimizing the MMD and using KFDA, consider a fixed kernel $k$ and denote by $\mathcal{A}$ the set of bounded positive operators on $\mathcal{H}_k$. We consider the nonparametric class of kernels $\mathcal{K} = \{{k}_A | k_A(x, y) = \braket{A k(x, \cdot), A k(y, \cdot)}, A \in \mathcal{A}\}$. For this class of kernels, we show in App.~\ref{app:nonparametric} that using \textsc{opt-mmd-witness} leads to the same witness function as using \textsc{kfda-witness}.

\paragraph{\textsc{kfda-boot}.}
It turns out that KFDA-like test statistics were considered before \citep{MouBacHar2008}, but in settings without data splitting.
Indeed, for a fixed $k$ and $\lambda > 0$, we can use the whole data, i.e., $\xfull, \yfull$ for learning the witness $ (\hat{\Sigma} + \lambda)^{-1} (\mu_{\xfull} - \mu_{\yfull})$ and computing the test statistic (empirical mean difference). The test statistic thus is $
    \tau_\textsc{kfda-boot} = \braket{\mu_{\xfull} - \mu_{\yfull}, (\hat{\Sigma} + \lambda)^{-1} (\mu_{\xfull} - \mu_{\yfull})},$
and we call its population version $\text{KFDA}^2 (P,Q|k, \lambda)$.
This, is the test statistic as studied by \citet{MouBacHar2008}.  As for \textsc{mmd-boot}, the same data is used for estimating the witness and computing the mean difference, hence Theorem \ref{thm:asympt_1d} does not hold anymore. We thus need to bootstrap the null distribution via permutations of the samples; thus, we refer to it as \textsc{kfda-boot}.
\textsc{kfda-boot} has similar drawbacks as $\textsc{mmd-boot}$:
1. simulating the null distribution via permutations has cost $\mathcal{O}((n+m)^3 B)$ for $B\in \mathbb{N}$ draws from the null distribution; and
2. we have to fix $k$ and $\lambda$ a priori, and their choices strongly affect the test power. \citet{MouBacHar2008} do not provide guidance for how to choose $k$ and $\lambda$.

\begin{table*}[t]
\caption{Overview of kernel-based two-sample tests. \textsc{a priori} means that the kernel/regularization is chosen independently of the data. The present work proposes the "witness" methods.}\label{table:kernel-methods}
\label{sample-table}
\begin{center}
\begin{small}
\begin{sc}
\resizebox{\textwidth}{!}{
\begin{tabular}{l|cccccc}
\toprule
Method & kernel choice & reg.~$\lambda$ & witness obj. & witness estim. & test data & threshold \\
\midrule
\textbf{\textsc{kfda-witness}}{\tiny (proposed)}  & CV & CV & SNR & $\ztr$ & $\zte$ & analytic \\
\textsc{kfda-boot}{\tiny \citep{MouBacHar2008}}  & a priori & a priori & SNR & $\zfull$ (implicit) & $\zfull$ & bootstrap \\
\textsc{mmd-boot}{\tiny \citep{gretton2012kernel}}  & a priori & - & MMD & $\zfull$ (implicit) & $\zfull$ & bootstrap \\
\textbf{\textsc{opt-mmd-witness}}{\tiny (proposed)}   & $J$ with $\ztr$  & - & MMD & $\ztr$ & $\zte$ & analytic \\
\textsc{opt-mmd-boot}{\tiny \citep{sutherland2016generative}}  & $J$ with $\ztr$  & - & MMD & $\zte$ (implicit) & $\zte$ & bootstrap \\
\bottomrule
\end{tabular}}
\end{sc}
\end{small}
\end{center}
\vskip -0.1in
\end{table*}

\section{RELATED WORK}\label{sec:related_work}
Besides the kernel-based tests we discussed so far, \citet{ChwialkowskiRSG15}
proposed tests based on \emph{smooth characteristic functions} (SCF), and
projected \emph{mean embeddings} (ME) of the distributions
where the mean embeddings are projected to $J$-dimensional Euclidean vectors for
$J \in \mathbb{N}$.
In fact, the normalized ME statistic in \citep[Eq.~13]{ChwialkowskiRSG15} can be
seen as a variant of the KFDA where the function classes is restricted by the $J$
projection directions.
Note that for a finite-dimensional RKHS and without regularization,
\textsc{kfda-boot} corresponds to the Hotelling's $T^2$ statistic
\citep{hotelling1931generalization}. \citet{JitSzaChwGre2016} improve this
approach by optimizing the features in the first stage. However, they also
discard the training data after learning the $J$ projection directions.
\citet{KirchlerKKL20} propose to learn a deep finite-dimensional representation of the data and to use this for a subsequent MMD or KFDA test. However, their  training objective does not directly maximize the test power \citep[Sec.~3.1.1]{KirchlerKKL20}.
\citet{liu2020learning} propose a deep version of \textsc{opt-mmd-boot}. They learn a deep-kernel (\textsc{mmd-d}) of the form $k_\omega (x,x') = \left[(1-\epsilon) \kappa(\phi_\omega(x),\phi_\omega(x'))) + \epsilon \right]q(x,x')$,
where $\epsilon \in (0,1)$, $\kappa$ and $q$ are Gaussian kernels and $\phi_\omega$ is a deep representation optimized via the criterion $J$, see App.~\ref{proof:u_stat_mmd_snr}. They also consider a version called \textsc{mmd-o} which is $k_\omega (x,x') =  \kappa(\phi_\omega(x),\phi_\omega(x'))$ and conclude that learning a full kernel (they advocate \textsc{mmd-d}) is better than learning a one-dimensional representation.

Most of the aforementioned works focus on developing a practical testing procedure for a specific dataset at hand. However, there also exist more theoretical work on the statistical optimality of different kernel-based approaches. \citet{Balasubramanian2021optimality} show that  a \emph{moderated} MMD  approach (which is related to KFDA) leads to optimal rates when testing against local alternatives. A similar discussion can be found in the long version of \citet[Sec.5.1]{harchaoui2008testing-long}. This resonates our findings, that a witness based on KFDA is more powerful than simply using the MMD witness. Furthermore, \citet{li2019optimality} show how the choice of scaling parameter in Gaussian kernels affects the statistical optimality. However, such theoretically optimal tests oftentimes are unpractical to use. \citet{Balasubramanian2021optimality}, for examples requires, the eigendecomposition of the kernel function, which generally is hard to obtain. Furthermore, without data splitting also these works cannot find a good kernel function.

Since our proposed witness function is one-dimensional, it is closely related to classification based two-sample tests \citep{Friedman03:Tests, kim2016classification, LopOqu2017, CheClo2019, CaiGogJia2020}. \citet{LopOqu2017} proposed learning a deep classifier and using its classification accuracy as test statistic. We refer to this as \textsc{c2st-s}, where \textsc{s} stands for sign. The method has two drawbacks. First, classification loss does optimize the 0-1 loss, whereas we directly maximize test power \citep[Remark 2]{LopOqu2017}. Second, it only uses the sign of the classification function and thus neglects information by weighting all points equally.  \citet{CheClo2019} address the second issue by considering the network's output before thresholding the function into a classifier. They train with a softmax loss, which also does not directly address test power. The connections of these methods to kernel-based tests were also thoroughly discussed by \citet{liu2020learning} and, in accordance, we refer to the approach of \citet{CheClo2019} as \textsc{c2st-l}.
\section{EXPERIMENTS}\label{sec:experiments}

\begin{figure*}[t]
    \centering
    \includegraphics[width=\linewidth]{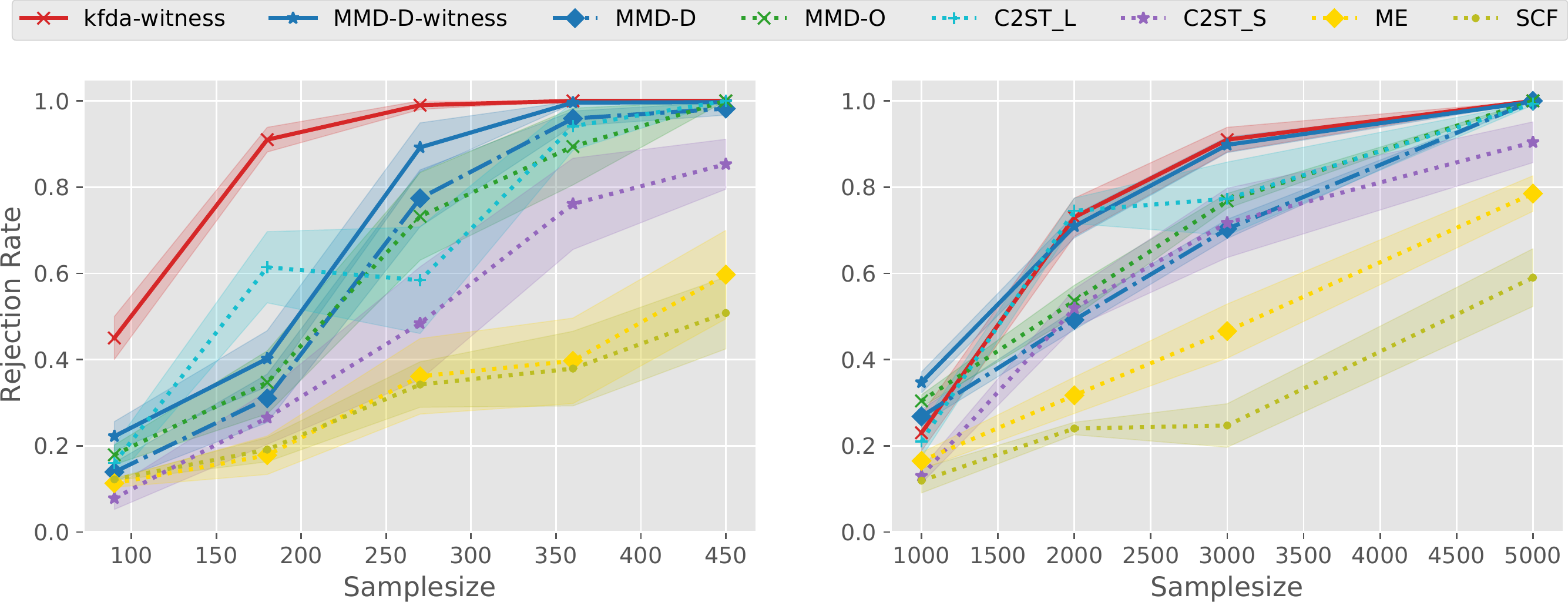}
    \caption{Benchmark experiments adapted from \citet{liu2020learning} \textbf{Left:} Blobs, \textbf{Right:} HIGGS. Computing the MMD witness after kernel optimization and performing a witness test (\textsc{mmd-d-witness}) improves the test power over \textsc{mmd-d}. Directly learning the \textsc{kfda-witness} also leads to high power. }
    \label{fig:benchmark}
\end{figure*}

We empirically assess the test power of the proposed WiTS tests in two settings. First, we perform instructive experiments to highlight the differences of the methods summarized in Table \ref{table:kernel-methods}. Second, we perform benchmark experiments on two challenging datasets and compare the performance of the introduced WiTS tests (\textsc{kfda-witness} and \textsc{opt-mmd-witness}) to the benchmarks (\textsc{mmd-d}, \textsc{mmd-o}, \textsc{me}, \textsc{scf}, \textsc{c2st-s}, \textsc{c2st-l}) introduced in Section \ref{sec:related_work}. For the benchmarks, we reuse the implementation provided by \citet{liu2020learning} without changing any hyperparameters.
Throughout our experiments we set the level $\alpha = 0.05$. App.~\ref{app:exp} contains experiments for correct type-I error control.
The shaded regions contain $\pm$ one standard deviation of the estimates.\footnote{Note that in Fig.~\ref{fig:benchmark} we used different approaches to estimate the rejection rates, see Appendix \ref{app:exp}. This explains that at the same rejection rate we can have differently large errors.}

\paragraph{Instructive experiments.}
In Figure \ref{Fig:instructive}, we consider  a \textbf{Blobs} dataset \citep{Gretton2012optimal} where $P$ and $Q$ are mixtures of nine anisotropic 2-d Gaussians with $Q$ having the covariance matrix rotated by an angle $\theta=\pi/4$, see Figure \ref{fig:blobs_draws} in the appendix.
For the left panel of Fig.~\ref{Fig:instructive}, we consider a single Gaussian kernel $k_\sigma(x, x') = \exp{(-\lVert x - x' \rVert^2/\sigma^2})$ with bandwidth $\sigma = 0.2$ and a  regularization parameter for the \textsc{kfda} methods of $\lambda = 10^{-2}$ (in the appendix we show the effect of the regularization in Fig.~\ref{Fig:kfda-regularization}. Note that for $\lambda \to \infty$, \textsc{kfda} and \textsc{mmd} methods coincide). We showcase the effect of varying splitting ratios $r$ when the kernel is fixed a-priori (thus we can apply \textsc{mmd-boot} and \textsc{kfda-boot}). With fixed kernel, \textsc{opt-mmd-boot} essentially discards the training data. We estimate the test power (rejection rate) with fixed overall sample size $n=m=100$.
We observe that the witness methods achieve highest power for a 50/50 split, given a fixed kernel and fixed regularization. We also observe that the boot approaches outperform the witness methods in this case.

However, in practice, it is unlikely that we can pick a powerful kernel and regularization \emph{a priori}. Therefore, for the middle panel of Figure \ref{Fig:instructive}, we optimize the kernel function over a class of kernels $\mathcal{K}_{10}$ consisting of ten Gaussian kernels with bandwidths on a logarithmic range from $10^{-3}$ to $10^1$. Additionally, for \textsc{kfda-witness} we cross-validate over five candidate regularizations on a log range from $10^{-4}$ to $10^3$. In this case, the witness methods attain the highest power at a splitting ratio $r>1/2$, and \textsc{opt-mmd-witness} outperforms \textsc{opt-mmd-boot} for the majority of splitting ratios and also globally.
For the right panel, we use the same setting, but fix the splitting ratio at $r=1/2$ and vary the sample size. As we expect, all tests are consistent and we observe that both WiTS test approaches outperform \textsc{opt-mmd-boot} at a 50/50 split.

\paragraph{Benchmark Experiments.} \citet{liu2020learning} benchmarked several deep classification two-sample tests (\textsc{c2st-l}, \textsc{c2st-c}) against MMD with an optimized deep kernel (\textsc{mmd-d}, \textsc{mmd-o}) and the optimized tests (\textsc{me}, \textsc{scf}) of \citet{JitSzaChwGre2016}. We implement \textsc{opt-mmd-witness} on top of their proposed method \textsc{mmd-d}, which optimizes a deep kernel \citep[Section 5]{liu2020learning}. Therefore after the kernel optimization, we use the training data to define the MMD witness function (Eq.~\eqref{eq:opt-mmd-witness}) and then proceed with \textsc{WitnessTest} from Algorithm \ref{alg:kfda_witness_test}. We also run \textsc{kfda-witness} with grid search over the same kernels and regularization as for the previous experiments.  We run the experiments on two benchmarks.
First, an adopted \textbf{Blobs} problem, with multiple different covariances \citep[Figure 1]{liu2020learning} (see Figure \ref{fig:blobs_draws} in the appendix), introduced to show the limitations of MMD with translation-invariant kernels. Second, the \textbf{Higgs} dataset \citep{baldi2014searching} where "we compare the jet $\phi$-momenta distribution ($d=4$) of the background process, $P$, which lacks Higgs bosons, to the corresponding distribution $Q$ for the process that produces Higgs bosons" (cited from \citet{liu2020learning}). For the Higgs dataset we consider sample sizes larger than a thousand per class.
To speed up the computation of the \textsc{kfda-witness},
we approximate the solution with $M=500$ Nyström centers, see Appendix \ref{app:KFDA_Falkon}, which underlines the scalability of our approach.
For both datasets we observe higher power of the WiTS tests we propose, see Figure \ref{fig:benchmark}. We emphasize that we used the implementation of \citet{liu2020learning}, without changing the deep architecture or any hyperparameters.

\section{CONCLUSION}\label{sec:conclusion}
We introduced a principled approach to learn optimal witness functions for two-sample testing.
The approach consists of two-stages: First, we learn a witness on a subset of the observations by maximizing a test-power criterion. In the second stage, we simply test whether the witness function attains the same mean on the test samples, and efficiently simulate the null distribution via permutations. 
We further showed how to adopt recent tests based on optimized Maximum Mean Discrepancy into a witness two-sample test.
\citet{liu2020learning} advocated optimizing a (deep) kernel in the training stage.
Our experiments show, however, that explicitly learning a one-dimensional witness can perform better than learning a high-dimensional representation (a kernel function) in the training stage.

Our results extend beyond kernel methods since we derive a principled objective to train a one-dimensional function optimal for two-sample testing. This objective and the proposed testing procedure can be applied with any function class.
The proposed framework thus not only allows domain experts to perform two-sample tests with the models most suitable to the data at hand, but can also easily incorporate model selection techniques developed for classification and regression tasks to optimize for the best parameter settings.
\paragraph{Acknowledgments} This work was in part supported by the German Federal Ministry of Education and Research (BMBF) through the Tübingen AI Center (FKZ: 01IS18039B) and the Machine Learning Cluster of Excellence number 2064/1 – Project 390727645.

\bibliography{refs}
\bibliographystyle{abbrvnat}

\newpage

\clearpage

\appendix
\onecolumn

\section{PROOFS}\label{app:proofs}

\subsection{Proof of Theorem \ref{thm:asympt_1d}}\label{app:proof_normality}
\begin{proof}
Theorem \ref{thm:asympt_1d} follows by the application of the CLT; see, e.g., Theorem A, Chapter 1.9.1 in \citet{serfling1980approximation}. The CLT implies $\sqrt{n+m} ( \hat{h}^n_P - \bar{h}_P ) = \sqrt{n/c} ( \hat{h}^n_P - \bar{h}_P ) \overset{d}{\to} \mathcal{N}(0, \sigma^2_P/c)$, analogously for $Q$ and the variances add up. Since $\hat\sigma^2_c(h) \overset{p}{\to} \sigma_c := \sigma^2_P / c + \sigma^2_Q/ (1-c)$, the result follows from Slutsky's theorem.
\end{proof}

\subsection{Proof of Proposition \ref{prop:consistency}}
\begin{proof}
Since we assume $\sigma_c(h)>0$, it follows that
\begin{align}
    \lim_{\nte+\mte \to \infty}\Phi\left( \Phi^{-1}(1-\alpha) - \sqrt{\nte+\mte} \frac{\bar{h}_P - \bar{h}_Q}{\sigma_c(h)}\right) = 0,
\end{align}
 i.e., the asymptotic rate of type-II errors goes to zero, if and only if $\bar{h}_P > \bar{h}_Q$.
\end{proof}
\subsection{Derivation of Equation \eqref{eq:witness_solution}}\label{app:witness_solution}
We use the following definitions: Let $Z = \{x_1,\dots,x_{\ntr}, y_1, \dots, y_{\mtr}\}$ denote the pooled training sample and $K$ denote the kernel matrix such that $K_{ij} = k(z_i,z_j)$ for $i,j \in [\ntr+\mtr]$. Let us define $G \in \mathcal{H}^{\ntr+\mtr}$ such that $G_i = k(z_i, \cdot)$. And we write $K = G^\top G$. Further we define $v_1 = (\frac{1}{\ntr}, \dots, \frac{1}{\ntr},0,\dots, 0)^\top\in \mathbb{R}^{\ntr+\mtr}$, $v_2 = (0, \dots, 0, \frac{1}{\mtr}, \dots, \frac{1}{\mtr})^\top\in \mathbb{R}^{\ntr+\mtr}$, and $\delta = v_1 - v_2$. For $l=\ntr,\mtr$ we define the idempotent centering operator $P_l = \mathop{I}_l - l^{-1} \bm{1}_l\bm{1}_l^\top$, where $\mathop{I}$ denotes the identity operator and $\bm{1}_l$ the $l$ dimensional vector with all ones. With this we define the $(\ntr+\mtr) \times (\ntr+ \mtr) $ matrix
$
    N_c = \begin{pmatrix}
    \frac{1}{c} P_{\ntr} & 0\\ 0 & \frac{1}{1-c}P_{\mtr}
    \end{pmatrix}.$
With the preceding definitions, we obtain $\hat{\mu}_P - \hat{\mu}_Q = G \delta$, $\hat{\Sigma} = \frac{1}{\ntr+\mtr} G N_c G^\top$.

Starting from \eqref{eq:reg_witness} we estimate the KFDA witness based on the empirical estimates of $\mu_P, \mu_Q, \Sigma$, i.e.,
\begin{align}\label{eq:empirical_estimate_primal}
    \hat{h}_\lambda = \argmax_{\substack{f \in \mathcal{H}}} \frac{\braket{\mu_{\xtr} - \mu_{\ytr}, f}}{\braket{f, (\hat\Sigma + \lambda \mathop{I}) f}^\frac{1}{2}}.
\end{align}
We first show a \emph{representer Theorem} for KFDA \citep[Sec.~3.4.3]{Mika_PhD2003}. Therefore, we decompose possible candidate functions $f=f_1+f_2 \in \mathcal{H}$ into a part $f_1$ that lies in the span of the training data $\mathcal{S}_{\text{tr}}=\text{span}(\{k(z_i, \cdot)|i \in [\ntr+\mtr]\})$ and $f_2$ which lies in the span's orthogonal complement.
Thus, by definition, we have $\braket{f_2, k(z_i, \cdot)}=0$ for all $i\in [\ntr+\mtr]$.
Since $\mu_{\xtr}$ and $\mu_{\ytr}$ are within $\mathcal{S}_\text{tr}$, we have $\braket{\mu_{\xtr} - \mu_{\ytr}, f} = \braket{\mu_{\xtr} - \mu_{\ytr}, f_1}$. Similarly, since $\hat{\Sigma}$ is only defined via the training samples in $Z$, $\hat{\Sigma}$ maps functions from $\mathcal{S}_\text{tr} $ to $\mathcal{S}_\text{tr}$ and we have $\hat{\Sigma} f_2 = 0$. Thus for the denominator of \eqref{eq:empirical_estimate_primal} we get
\begin{align}
    \braket{f, (\hat\Sigma + \lambda \mathop{I}) f} =  \braket{f_1, (\hat\Sigma + \lambda \mathop{I}) f_1} + \lambda \|f_2\|^2 \geq \braket{f_1, (\hat\Sigma + \lambda \mathop{I}) f_1}.
\end{align}
We have shown that the nominator of \eqref{eq:empirical_estimate_primal} stays constant, if we add a function $f_2$ that is not is not in $\mathcal{S}_\text{tr}$ and the denominator can only grow. This implies that the maximum in \eqref{eq:empirical_estimate_primal} is attained for a function in $\mathcal{S}_\text{tr}$ and we can expand it as
$\hat{h}_\lambda(\cdot) = \sum_{i=1}^{\ntr+\mtr} \hat{\alpha}_i k(z_i, \cdot)$. Hence the solution is
\begin{align}\label{eq:maximization_dual}
    \hat{\alpha} &= \argmax_{\alpha \in \mathbb{R}^{\ntr+\mtr}} \frac{\braket{\mu_{\xtr} - \mu_{\ytr}, \sum_{i=1}^{\ntr+\mtr} \alpha_i k(z_i, \cdot)}}{\braket{\sum_{i=1}^{\ntr+\mtr} \alpha_i k(z_i, \cdot), (\hat{\Sigma} + \lambda \mathop{I}) \sum_{i=1}^{\ntr+\mtr} \alpha_i k(z_i, \cdot)}^\frac{1}{2}} \\
    &= \argmax_{\alpha \in \mathbb{R}^{\ntr+\mtr}} \frac{\delta^\top K \alpha}{\left(\alpha^\top \left(\frac{KN_cK}{\ntr +\mtr} + \lambda K\right)\alpha\right)^\frac{1}{2}}.
\end{align}
The solution to this is \citep[Sec.~3.2]{Mika_PhD2003}\footnote{For a sanity check, simply compute the gradient of \eqref{eq:maximization_dual} and set it to zero.}
\begin{align}
    \left(\frac{KN_cK}{\ntr +\mtr} + \lambda K\right)\hat{\alpha} = K\delta \qquad \Longleftrightarrow \qquad \hat{\alpha} =\left(\frac{KN_cK}{\ntr +\mtr} + \lambda K\right)^{-1} K\delta.
\end{align}

\subsection{Convergence of $\hat{h}_\lambda$}\label{app:convergence}
We will show that $\hat{h}_\lambda \to h_\lambda = (\Sigma + \lambda \mathop{I})^{-1} (\mu_P - \mu_Q)$ in probability.
\begin{proof}
First, we observe that
\begin{align*}
    \hat{h}_\lambda - h_\lambda &= (\hat{\Sigma} +\lambda I)^{-1} (\mu_{\xtr} - \mu_{\ytr}) - ({\Sigma} +\lambda I)^{-1} (\mu_P - \mu_Q)\\
    &= (\hat{\Sigma} +\lambda I)^{-1} (\mu_{\xtr} - \mu_{\ytr}) - (\hat{\Sigma} +\lambda I)^{-1} (\mu_P - \mu_Q) \\
    & \qquad +(\hat{\Sigma} +\lambda I)^{-1} (\mu_P - \mu_Q) - ({\Sigma} +\lambda I)^{-1} (\mu_P - \mu_Q)\\
    &=(\hat{\Sigma} +\lambda I)^{-1} \left[(\mu_{\xtr} - \mu_{\ytr}) - (\mu_P - \mu_Q)\right] +\left[(\hat{\Sigma} +\lambda I)^{-1}  - ({\Sigma} +\lambda I)^{-1} \right](\mu_P - \mu_Q).
\end{align*}
Thus it follows that
\begin{align*}
    \lVert\hat{h}_\lambda - h_\lambda \rVert_\mathcal{H} &\leq \lVert(\hat{\Sigma} +\lambda \mathop{I})^{-1} [(\mu_{\xtr} - \mu_{\ytr}) - (\mu_P - \mu_Q)] \rVert_\mathcal{H}  \\
    & \qquad + \lVert [(\hat{\Sigma} +\lambda \mathop{I})^{-1}  - ({\Sigma} +\lambda \mathop{I})^{-1} ](\mu_P - \mu_Q)\rVert_\mathcal{H} \\
    & = (A) + (B).
\end{align*}

\paragraph{Probabilistic bound on $(A)$.}
By the triangle inequality,
\begin{align*}
    \lVert(\hat{\Sigma} +\lambda \mathop{I})^{-1} [(\mu_{\xtr} - \mu_{\ytr}) - (\mu_P - \mu_Q)] \rVert_\mathcal{H} &\leq
    \lVert(\hat{\Sigma} +\lambda \mathop{I})^{-1}\rVert \lVert (\mu_{\xtr} - \mu_{\ytr}) - (\mu_P - \mu_Q) \rVert_\mathcal{H} \\
    &\leq \lVert(\hat{\Sigma} +\lambda \mathop{I})^{-1}\rVert (\lVert \mu_{\xtr}  - \mu_P\rVert_\mathcal{H} + \lVert\mu_Q - \mu_{\ytr}\rVert_\mathcal{H}).
\end{align*}
By the spectral theorem, $\|(\hat{\Sigma} + \lambda I)^{-1}\| = \sup_{\hat{l}\in(\hat{l}_k)_{k=1}^\infty}\frac{1}{\hat{l} +\lambda} \leq 1/\lambda$ where $(\hat{l}_k)_{k=0}^\infty$ are the eigenvalues of $\hat{\Sigma}$ and by definition non-negative.
Then, the $\sqrt{n}$-convergence of $(A)$ follows from the $\sqrt{n}$-convergence of the kernel mean embeddings $\left\lVert\mu_{\xtr} - \mu_P\right\rVert_{\mathcal{H}} = \mathcal{O}_p(\ntr^{-1/2})$ and $\left\lVert \mu_Q - \mu_{\ytr} \right\rVert_{\mathcal{H}} = \mathcal{O}_p(\mtr^{-1/2})$; see, e.g., \citet[Theorem 3.4]{Muandet2017}. That is, $(A) = \mathcal{O}_p(\min(\ntr,\mtr)^{-1/2})$.

\paragraph{Probabilistic bound on $(B)$.}
Using the identity $C^{-1} - D^{-1} = C^{-1}(D - C)D^{-1}$, we can rewrite $(B)$ as
\begin{align*}
    \lVert [(\hat{\Sigma} &+\lambda \mathop{I})^{-1}  - ({\Sigma} +\lambda \mathop{I})^{-1} ](\mu_P - \mu_Q)\rVert_\mathcal{H}
    \\
    &=\lVert (\hat{\Sigma} +\lambda \mathop{I})^{-1}(\hat{\Sigma}   - \Sigma )({\Sigma} +\lambda \mathop{I})^{-1}(\mu_P - \mu_Q)\rVert_\mathcal{H}\\
    &\leq \lVert (\hat{\Sigma} +\lambda \mathop{I})^{-1} \rVert \lVert\hat{\Sigma}   - \Sigma \rVert \lVert({\Sigma} +\lambda \mathop{I})^{-1}(\mu_P - \mu_Q) \rVert_\mathcal{H}\\
    &\leq \lVert (\hat{\Sigma} +\lambda \mathop{I})^{-1} \rVert \lVert\hat{\Sigma}   - \Sigma \rVert_{\text{HS}} \lVert({\Sigma} +\lambda \mathop{I})^{-1}(\mu_P - \mu_Q) \rVert_\mathcal{H},
\end{align*}
where we used that the operator norm is upper bounded by the Hilbert-Schmidt norm.
Let $n := \ntr + \mtr$. Then, since $\lVert(\hat{\Sigma} +\lambda \mathop{I})^{-1}\rVert \leq 1/\lambda$, the $\sqrt{n}$-convergence of $(B)$ follows from the $\sqrt{n}$-convergence of the covariance operator, i.e., $\lVert\hat{\Sigma} - \Sigma\rVert_{\text{HS}} = \mathcal{O}_p(n^{-1/2})$ \citep[Lemma 4]{FukumizuBG05}.
That is, $(B) = \mathcal{O}_p((\ntr + \mtr)^{-1/2})$.

Combining the rates of $(A)$ and $(B)$ yields the overall rate of convergence: $\lVert\hat{h}_\lambda - h_\lambda\rVert_\mathcal{H} = \mathcal{O}_p (\min(\ntr,\mtr)^{-1/2})$.
\end{proof}

\subsection{Witness objective vs.~kernel optimization objective in MMD tests}\label{proof:u_stat_mmd_snr}
In MMD-based two sample tests, the most common estimate of the MMD is the U-statistic estimate, defined as \citep{gretton2012kernel}
\begin{align}\label{eq:U-stat_mmd}
    \widehat{ \text{MMD}^2_u} = \frac{1}{n(n+1)} \sum_{i\neq j} H_{ij},
\end{align}
with $H_{ij} = \braket{k(x_i, \cdot) - k(y_i, \cdot), k(x_j, \cdot) - k(y_j, \cdot)}$.
The objective function used in \citet{sutherland2016generative, liu2020learning} bases on the asymptotic variance of the estimator under the alternative hypothesis.
If the population value of $\text{MMD}^2$ is positive, then the distribution of the estimate is asymptotically normal \citep[Section 5.5.1]{serfling1980approximation},
$
    \sqrt{n} \left(\widehat{ \text{MMD}_u^2} - \text{MMD}^2\right) \overset{d}{\to} \mathcal{N}(0, \sigma^2_{H_1}),
$
with $\sigma^2_{H_1} = 4 (\expect{}{H_{12}H_{13}} - \expect{}{H_{12}}^2)$
\citep{liu2020learning}.
This can be used to derive an asymptotic test power criterion, which is given as the signal-to-noise ratio $J = \frac{\text{MMD}^2}{\sigma_{H_1}}$ \citep[Sec.~2.1]{sutherland2016generative}.

We show, that the power criterion $J = \frac{\text{MMD}^2}{\sigma_{H_1}}$ corresponds to the SNR criterion we derived in \eqref{eq:general_objective}.
It is an easy exercise to show that
\begin{align*}
        \sigma^2_{H_1} = 4 &\left(\expect{X\sim P}{\braket{\mu_P - \mu_Q, k(X, \cdot)}^2}
        + \expect{Y\sim Q}{\braket{\mu_P - \mu_Q, k(Y, \cdot)}^2} \right.\\
         & \left. \quad -\braket{\mu_P - \mu_Q, \mu_P}^2
         - \braket{\mu_P - \mu_Q, \mu_Q}^2)\right).
\end{align*}
    Recalling the definition of the covariance operator $\Sigma_P = \expect{}{k(X,\cdot)\otimes k(X,\cdot)} - \mu_P \otimes \mu_P$, we obtain
\begin{align*}
    \sigma^2_{H_1} &= 4\braket{\mu_P - \mu_Q, (\Sigma_P + \Sigma_Q)(\mu_P - \mu_Q)} = 2\braket{\mu_P - \mu_Q, (2\Sigma_P + 2\Sigma_Q)(\mu_P - \mu_Q)} \\
    &= 2\braket{\mu_P - \mu_Q, \Sigma(\mu_P - \mu_Q)},
\end{align*}
where we used $\Sigma = \Sigma_P/c + \Sigma_Q/(1-c)$ and $c=1/2$ for balanced samples.

Using $h_k^{P,Q} = \mu_P - \mu_Q$, we have
\begin{align}\label{eq:J_mmd_snr}
    J(P,Q|k) &= \frac{\text{MMD}^2}{\sigma_{H_1}} =  \frac{\braket{\mu_P - \mu_Q,\mu_P - \mu_Q}}{\sqrt{2}\braket{\mu_P - \mu_Q, \Sigma(\mu_P - \mu_Q)}^{\frac{1}{2}}} = \frac{\braket{\mu_P - \mu_Q, h_k^{P,Q}}}{\sqrt{2}\braket{h_k^{P,Q}, \Sigma h_k^{P,Q}}^{\frac{1}{2}}} \\
    &= \frac{1}{\sqrt{2}} \text{SNR}(h_k^{P,Q}).
\end{align}

\subsection{MMD of nonparametrically optimized kernel corresponds to KFDA}\label{app:nonparametric}
Consider a fixed kernel $k$ and denote by $\mathcal{A}$ the set of bounded positive operators on $\mathcal{H}_k$.
For the nonparametric class of kernels $\mathcal{K} = \{{k}_A | k_A(x, y) = \braket{A k(x, \cdot), A k(y, \cdot)}, A \in \mathcal{A}\}$ using \textsc{opt-mmd-witness} leads to exactly the same witness function as using \textsc{kfda-witness}.
\begin{proof}
Writing inner products in the original RKHS with kernel $k$ for kernel $k_A$ we have the regularized $J$ criterion
$$J_A^\lambda =  \frac{\braket{A(\mu_P - \mu_Q),A(\mu_P - \mu_Q})}{\braket{A(\mu_P - \mu_Q), A(\Sigma+\lambda \mathop{I})AA(\mu_P - \mu_Q)}^{\frac{1}{2}}}.$$
    We define $\delta_A := A^2 (\mu_P - \mu_Q)$ and obtain
\begin{align}
    J_A^\lambda = \frac{\braket{\mu_P - \mu_Q,\delta_A})}{\braket{\delta_A, (\Sigma+\lambda \mathop{I})\delta_A}^{\frac{1}{2}}},
\end{align}
which looks almost like \eqref{eq:reg_witness}. The solution to \eqref{eq:reg_witness} is \eqref{eq:reg_fisher_witness_explicit} which implies that $\tilde{A}_\lambda = (\Sigma + \lambda \mathop{I})^{-\frac{1}{2}}$ defines the optimal kernel
\begin{align*}
    \tilde{k}_\lambda(x,x') &:= \braket{(\Sigma +\lambda \mathop{1})^{-\frac{1}{2}}k(x,\cdot),(\Sigma +\lambda \mathop{1})^{-\frac{1}{2}}k(x',\cdot)}_\mathcal{H}\\
    &= \braket{k(x,\cdot),(\Sigma +\lambda \mathop{1})^{-1}k(x',\cdot)}_\mathcal{H}.
\end{align*}

Based on the empirical estimates the MMD witness of the optimized kernel would be (expressed in terms of the original kernel $k$)
\begin{align}
    h_{\tilde{k}_\lambda}^{\ztr} = (\hat{\Sigma} +\lambda \mathop{1})^{-1} (\mu_{\xtr} - \mu_{\ytr}) = \hat{h}_\lambda,
\end{align}
i.e., the witness of \textsc{opt-mmd-witness} coincides with the \textsc{kfda-witness} in the original RKHS.
\end{proof}

\section{FURTHER EXPERIMENTS AND DETAILS}\label{app:exp}
\begin{figure}[t]
    \centering
    \subfigure{\includegraphics[width=.32\linewidth]{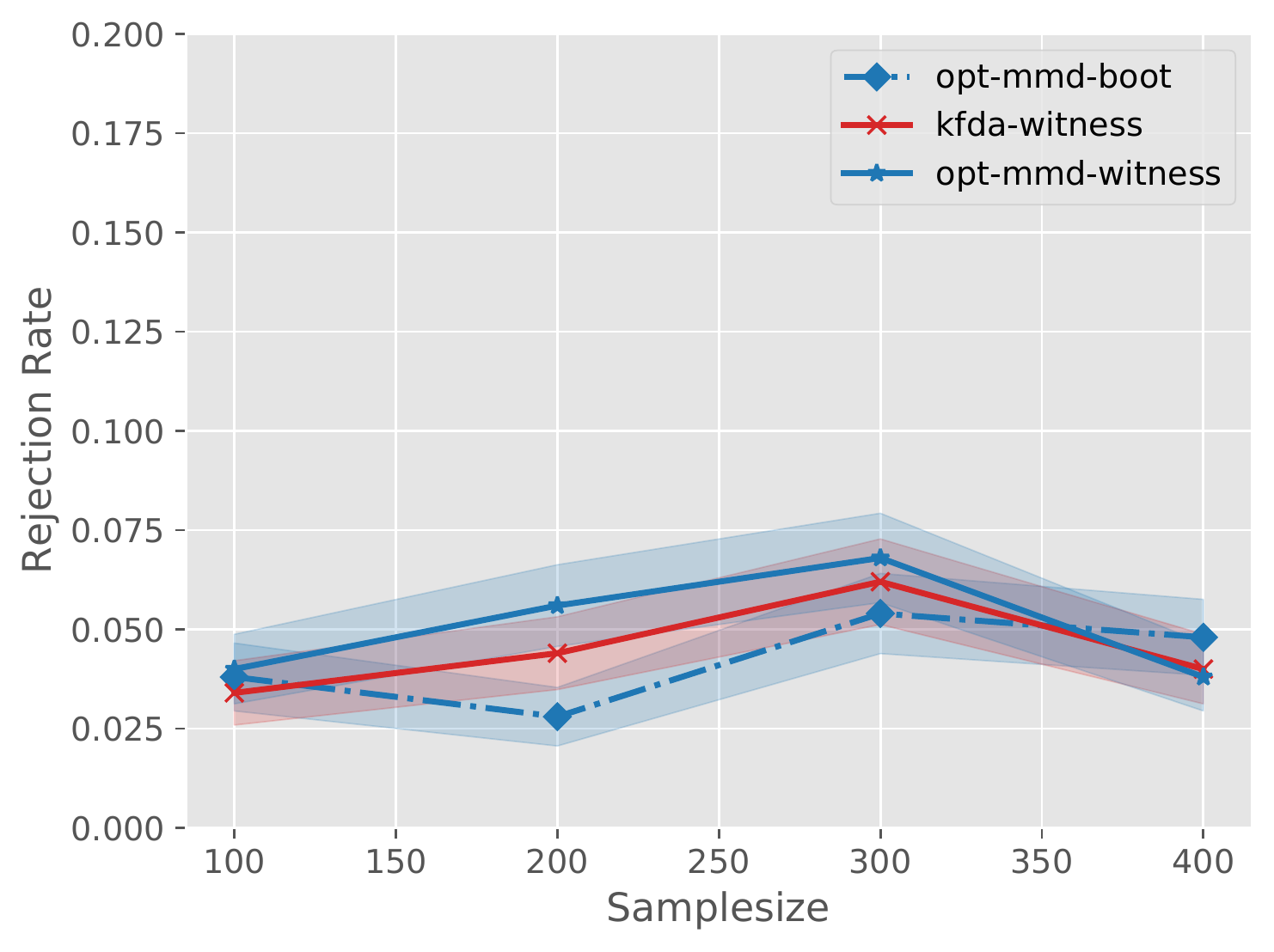}}
    \centering
    \subfigure{\includegraphics[width=.32\linewidth]{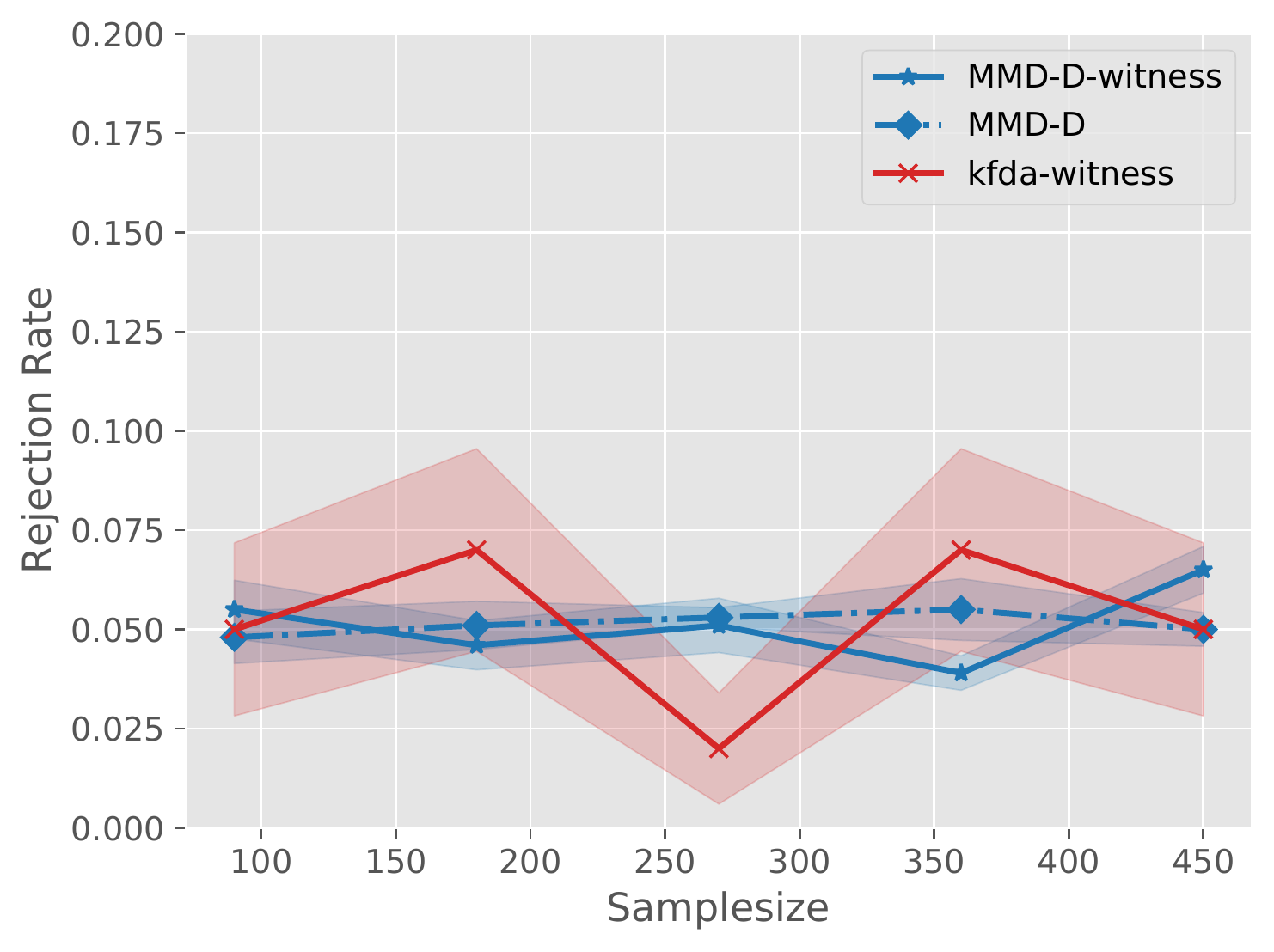}}
    \centering
    \subfigure{\includegraphics[width=.32\linewidth]{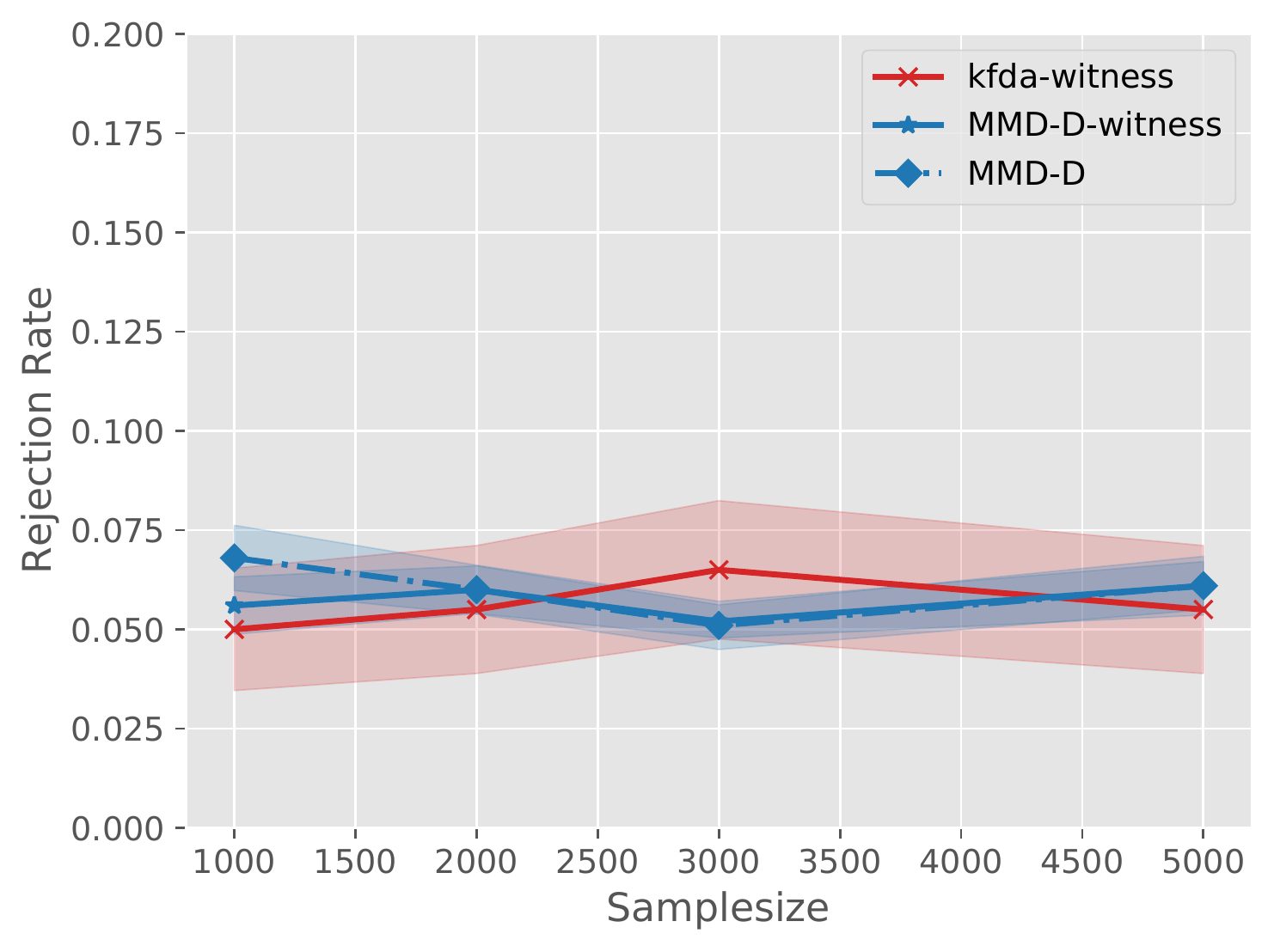}}
    \caption{Rejection Rates for true null hypothesis (Type I error) at $\alpha = 0.05$. \textbf{Left:} Standard Blobs dataset (500 iterations). \textbf{Middle:} Blobs dataset of \citet{liu2020learning}, \textsc{kfda-witness} is only average over 100 trials the others over $10\times 100$, therefore \textsc{kfda-witness} has higher variance. \textbf{Right:} Higgs dataset}
    \label{fig:exp1-typeI}
\end{figure}

\begin{figure*}[th]
    \centering
    \subfigure{\includegraphics[width=.32\linewidth]{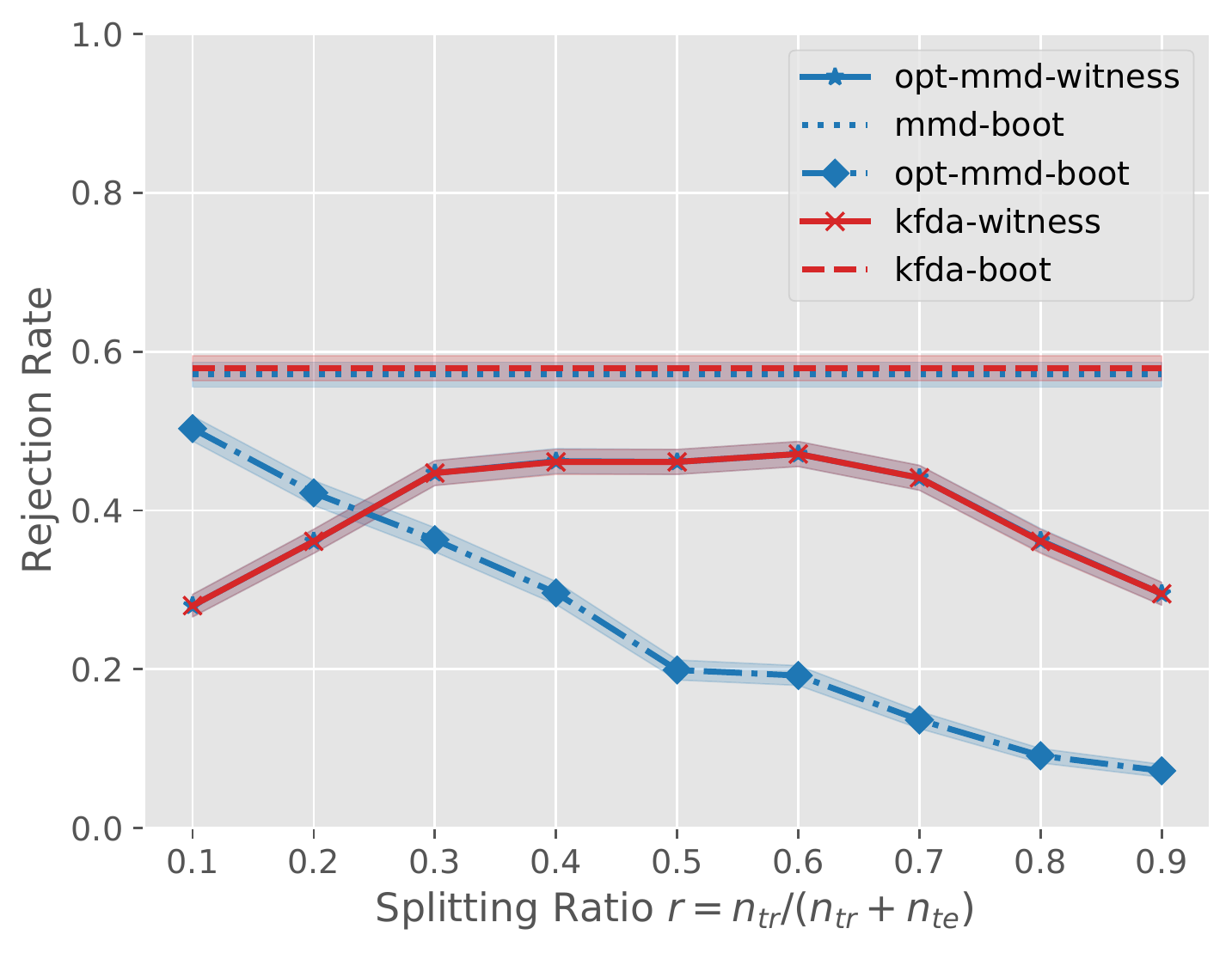}}
    \centering
    \subfigure{\includegraphics[width=.32\linewidth]{figures/a_exp_reg-e-2_error.pdf}}
    \centering
    \subfigure{\includegraphics[width=.32\linewidth]{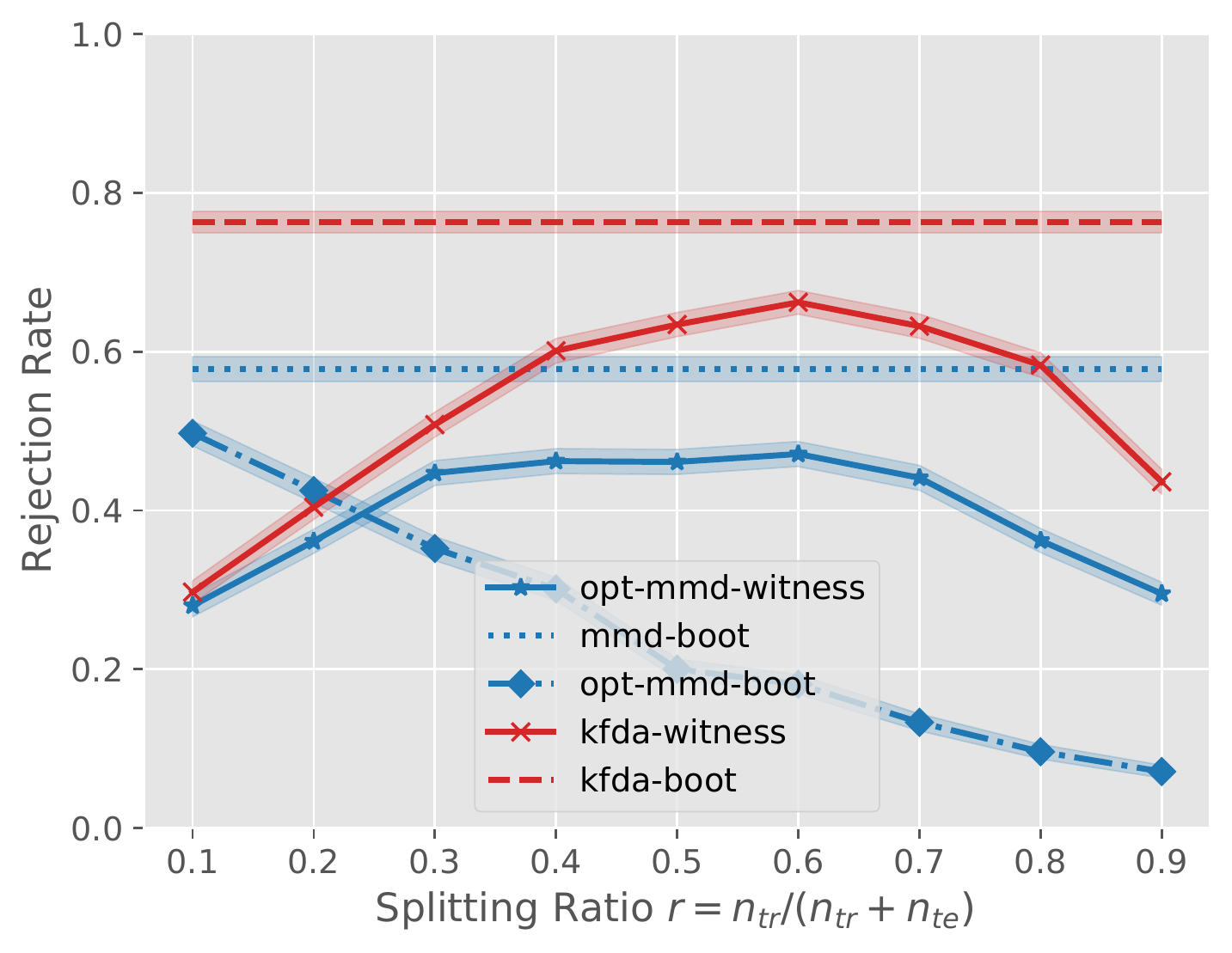}}
    \caption{\textbf{Effect of regularization on KFDA.} We consider the same setting as in the left panel of Fig.~\ref{Fig:instructive} (fixed kernel and fixed regularization and $n=m=100$) but for different regularization. \textbf{Left $(\lambda=10^3)$:} For large regularization KFDA converges to MMD. \textbf{Middle ($\lambda=10^{-2}$):} For a good regularization the KFDA approaches clearly outperform the corresponding MMD approaches. \textbf{Right $(\lambda=10^{-4}$):} If the regularization is too small for a given sample size (here $n=100$) , then KFDA overfits in the training phase, which leads to a reduction in test power. }
    \label{Fig:kfda-regularization}
\end{figure*}
\begin{figure}[t]
    \centering
    \subfigure{\includegraphics[width=.48\linewidth]{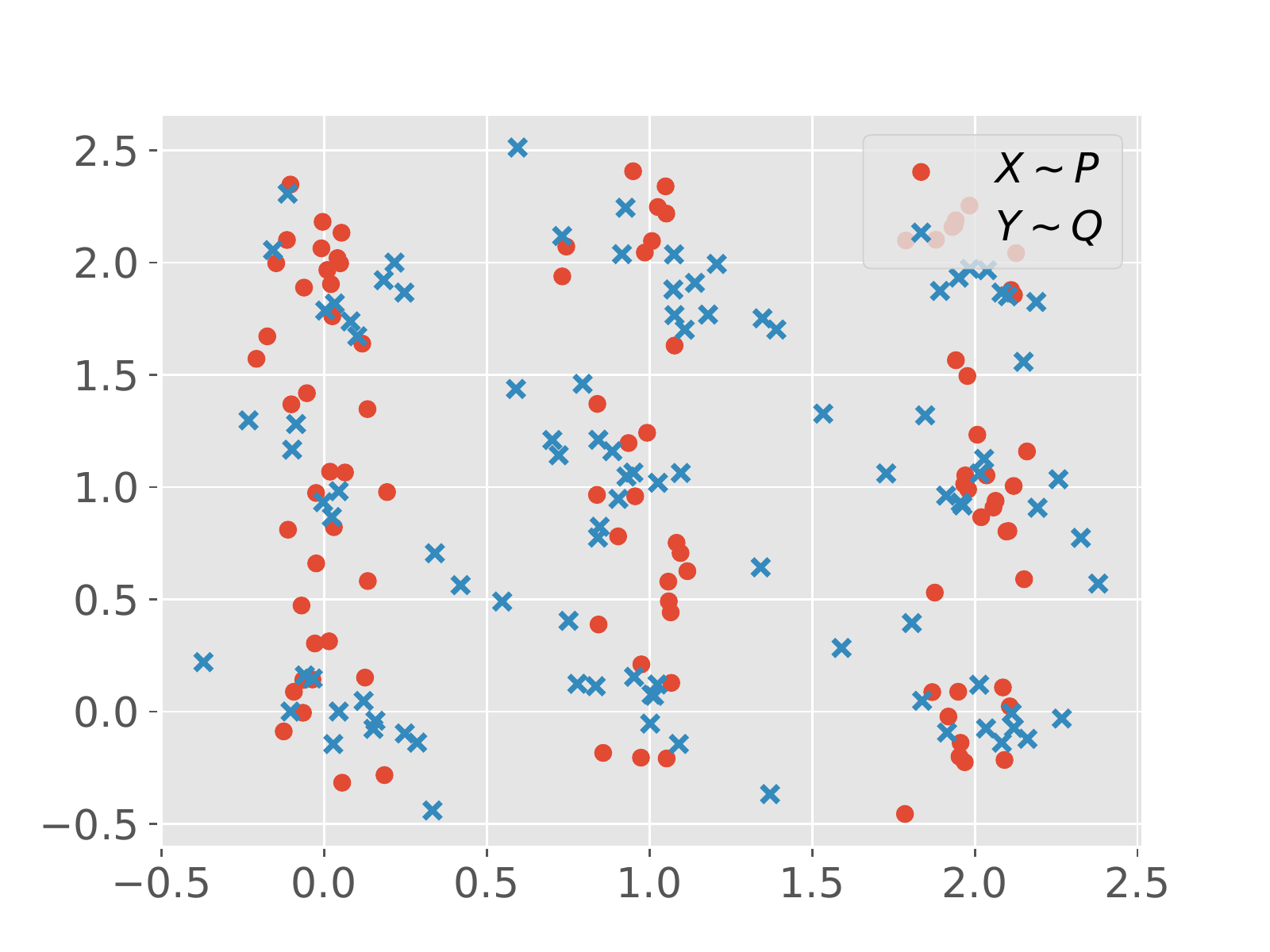}}
    \centering
    \subfigure{\includegraphics[width=.48\linewidth]{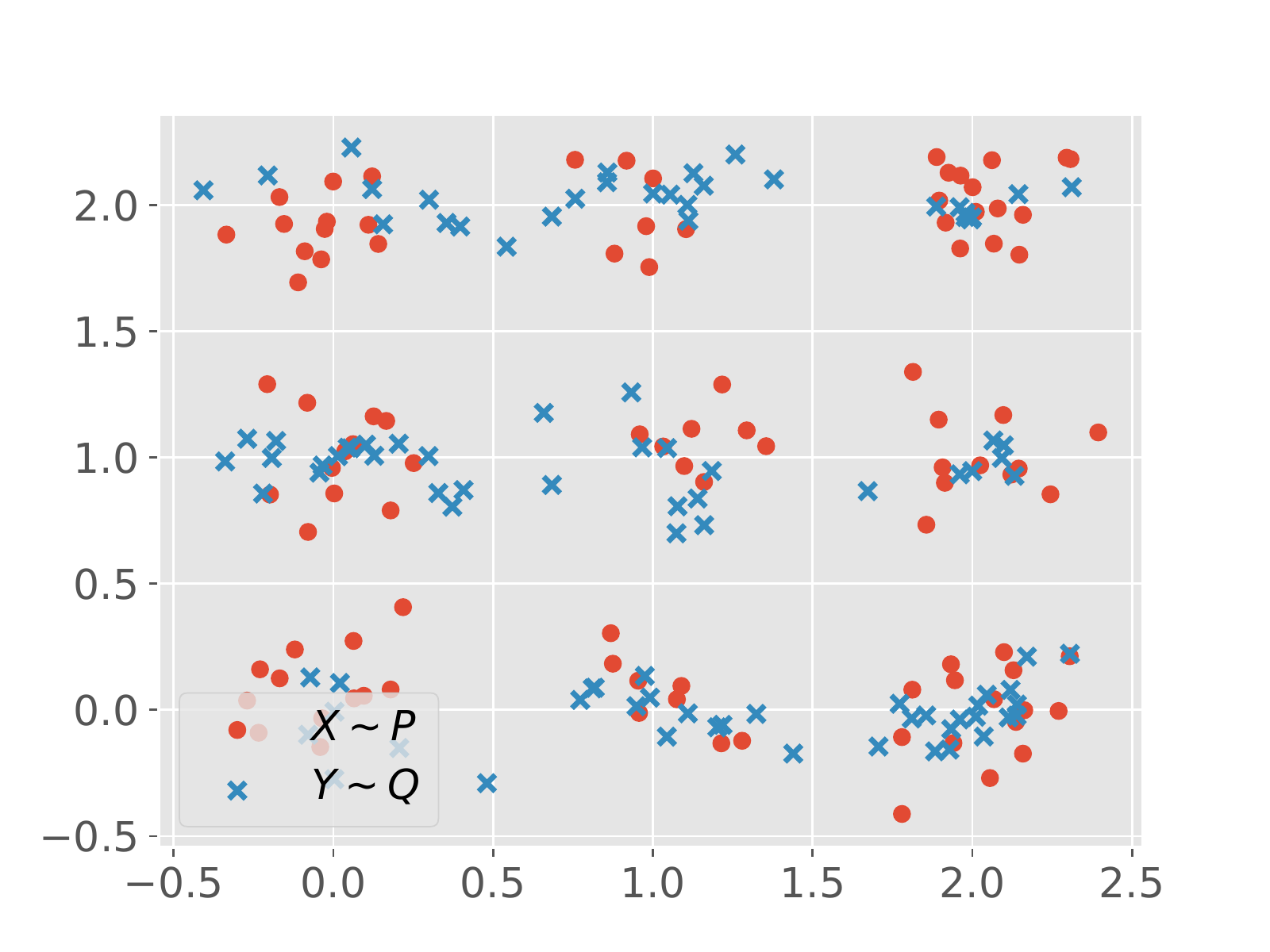}}

    \caption{\textbf{Left:} Draws from Blobs dataset for the instructive experiments. The distributions are mixtures of nine Gaussians, with anisotropic covariance (but the same covariance matrix across blobs). The covariance matrix of $Q$ is rotated by $\theta= \pi/4$ relative to the covariance matrix of $P$. To simulate the null hypothesis we use $\theta = 0$, which corresponds to drawing both samples from $P$. \textbf{Right:} Blobs dataset used for Figure \ref{fig:benchmark} as suggested by \citet[Figure 1]{liu2020learning}. In this case, $P$ has isotropic Gaussian, the blobs in $Q$ are anisotropic and have different covariance matrices. To simulate the null hypothesis, we draw both samples from $P$.}
    \label{fig:blobs_draws}
\end{figure}

This section provides supplementary information on our experiments. We provide code upon personal request.

\paragraph{Datasets.} We used two different versions of the Blobs dataset. We show random draws for both cases in Figure \ref{fig:blobs_draws}. For the benchmark experiments we also used the Higgs dataset \citep{baldi2014searching}, which is part of the \emph{UCI Machine Learning Repository} (\url{https://archive.ics.uci.edu/ml/datasets/HIGGS}). We used a version that is ready for Python usage provided by \citet{liu2020learning} (\url{https://drive.google.com/open?id=1sHIIFCoHbauk6Mkb6e8a_tp1qnvuUOCc}). To ensure the comparability we follow the implementation of \citet{liu2020learning} and draw samples from the Higgs dataset \emph{without replacement.}

\paragraph{Effect of regularization of \textsc{kfda-witness}.} In the left panel of Figure \ref{Fig:instructive}, we chose a fixed regularization $\lambda = 10^{-2}$ for the KFDA methods. In Figure \ref{Fig:kfda-regularization}, we show the effect of choosing a bad regularization. If the regularization is too large (left), then KFDA coincides with MMD. On the other hand, if the regularization is too small (right), then the effect of inaccurately estimating the covariance operator might as well lead to a reduced test power. For good performance it is thus important to chose a suitable regularization. This can be automated by including a model selection procedure, such as cross-validation, in the training stage.

\paragraph{Estimation of Rejection Rates.}
For the instructive experiments in Figure \ref{Fig:instructive} we estimate the rejection rates by repeating the whole two-stage procedure 1000 times.
For the benchmark experiments we use 100 iterations of the two-stage procedure for \textsc{kfda-witness}.
For all the other methods in the benchmark experiments, we follow the implementation of \citet{liu2020learning} and estimate the rejection rates by running the first stage ten times and estimating the rejection rate over 100 independent test sets for each run of the first stage. The reason for this is, that the first stage is quite slow (training a neural network).

\paragraph{Type-I errors.} We report Type-I errors for all three different datasets in Figure \ref{fig:exp1-typeI}.

\section{APPROXIMATE COMPUTATION OF THE KFDA WITNESS}
\label{app:KFDA_Falkon}
\begin{algorithm}[t]
	\caption{Pseudocode for the FdaFalkon algorithm. Adopted for KFDA from \citep{Meanti2020} \label{alg:falkon}}\label{Alg:KFDA_Implementation}
	\begin{minipage}{0.53\linewidth}{
			\footnotesize
			\begin{algorithmic}[1]
				\footnotesize
				\Function{FdaFalkon}{$Z, \bm{y}, k,  \lambda, m$, t}
				\State $Z_{m}, \bm{y}_m \gets$ \Call{RandomSubsample}{$(Z,\bm{y}), m$} %
				\State $T, A \gets $ \Call{Preconditioner}{$Z_m,\bm{y}_m, \lambda$}
				\Function{LinOp}{$\bm{\beta}$} %
				\State $\bm{v} \gets A^{-1}\bm{\beta}$
				\State $\bm{c} \gets k(Z_m, Z)N N^\top
				k(Z, Z_m)T^{-1}\bm{v}$
				\State \textbf{return} $A^{-\top} (T^{-\top} \bm{c} + \lambda n \bm{v})$
				\EndFunction

				\State $R \gets A^{-\top}T^{-\top}k(Z_m, Z)\bm{y}$ %
				\State $\bm{\beta} \gets $ \Call{ConjugateGradient}{$\textsc{LinOp}, R, t$} %
				\State \textbf{return} $T^{-1}A^{-1} \bm{\beta}$, $Z_m$ %
				\EndFunction
			\end{algorithmic}
		}
	\end{minipage}~~%
	\begin{minipage}{0.46\linewidth}
		\footnotesize
		\begin{algorithmic}[1]
			\setcounter{ALG@line}{12}
			\footnotesize
			\Function{Preconditioner}{$Z_m, \bm{y}_m, \lambda$}
			\State $K_{mm} \gets k(Z_m, Z_m)$  %
			\State $T \gets \chol(K_{mm})$ \label{line:chol1} %
			\State $K_{mm} \gets \frac{1}{m} TN_m N_m T^\top  + \lambda \eye$ \label{line:lauum_bg} %
			\State $A \gets \chol(K_{mm})$ \label{line:chol2}
			\State \textbf{return} $T, A$ %
			\EndFunction

			\Function{kfdaWitness}{$\ztr, k, \lambda$}
			\State $Z \gets$ \Call{Concatenate}{$\ztr$}
			\State $\bm{y} = [1]*\Call{len}{\xtr} + [-1]*\Call{len}{\ytr}$
			\label{line:kernel} %
			\State $m = \Call{len}{Z}$ \Comment{\# Nyström centers}
			\State $\alpha, Z \gets$ \Call{FdaFalkon}{$Z, \bm{y}, k, \lambda, m$}
			\State \textbf{return} $h_\lambda = \sum_{i=1}^m \alpha_i k(z_i, \cdot)$
			\EndFunction
		\end{algorithmic}
		\end{minipage}%
\end{algorithm}%
In this section we will use $n$ instead of $\ntr$ and $m$ instead of $\mtr$ to keep the notation more concise.
In \ref{app:witness_solution}, we showed that the exact solution for the estimate of the KFDA witness is given by
\begin{align}
    \hat{h}_\lambda(\cdot) &= \sum_{i=1}^{n+m} \hat{\alpha}_i k(z_i, \cdot),\\
    \hat{\alpha} &= \left(\frac{KN_cK}{n +m} + \lambda K\right)^{-1} K\delta.
\end{align}
\begin{remark}
    The problem with computing the KFDA witness is that a naive implementation scales cubically with the pooled sample size. In this section, we thus derive an approach that builds on recent results, that show that one can essentially get optimal convergence guarantees while only using $\mathcal{O}((n+m)^{3/2})$ time. Therefore two steps are needed. First, the solution is approximated with $M = \mathcal{O}((n+m)^\frac{1}{2})$ Nystrom centers. Second the solution with for the Nystrom centers is found via conjugate gradient, where a preconditioner is computed again with only $M$ datapoints.
\end{remark}
We take an approach similar to \citet{Rudi2017, Meanti2020}. We will thus explicitly assume that the function $h$ has the parametric form
\begin{align}
    h_{\tilde{\alpha}}(x) = \sum_{m=1}^M \tilde{\alpha}_i k(x, \tilde{z}_i),
\end{align}
with $M = \{\tilde{z}_1, \dots, \tilde{z}_M\} \subseteq \{x_1, \dots, x_n, y_1, \dots, y_m\}$ (we overload notation and use $M$ to denote the set itself as well as its size). We take the notation introduced in Section \ref{sec:kfda_witness} and constrain to the case $c=\frac{1} {2}$. In this case we can use $N = \begin{pmatrix}
    P_n & 0\\ 0 & P_m
    \end{pmatrix} = \frac{N_c}{2}$, instead of $N_c$. Note that this only affects the scaling of the solution (if we also scale $\lambda$ accordingly), which is unimportant for WiTS tests. Using $N$ instead of $N_c$ has the advantage that $N$ itself is idempotent $N = N N^\top$, which makes the following easier. Nevertheless, it is straightforward to use the below algorithm for any $c \in(0,1)$, simply by using $N_c = \begin{pmatrix}
    \frac{1}{\sqrt{c}} P_n & 0\\ 0 & \frac{1}{\sqrt{1-c}} P_m
    \end{pmatrix} \begin{pmatrix}
    \frac{1}{\sqrt{c}} P_n & 0\\ 0 & \frac{1}{\sqrt{1-c}} P_m
    \end{pmatrix} $.

    In the following we denote with $K_{ZM}$ the $(n+m)\times M$ matrix of entries $k(z_i, \tilde{z}_j)$ and $K_{MZ}$ its transpose.
We can then rewrite the terms in our objective
\begin{align}
    &\braket{\hat{\mu}_P - \hat{\mu}_Q, h_{\tilde{\alpha}}} = \delta^\top K_{ZM} \tilde{\alpha},\\
    &\begin{aligned}
           &\braket{h_{\tilde{\alpha}}, (\hat{\Sigma} + \lambda \mathop{1}) h_{\tilde{\alpha}}} \\
           &\;= \tilde{\alpha}^\top \left(\frac{1}{n+m} K_{MZ}N N^\top K_{ZM} + \lambda K_{MM} \right) \tilde{\alpha}.
    \end{aligned}
\end{align}
Let us define $R_{MZ} := K_{MZ}N$. This is a $M \times (n+m)$ matrix. Note that $N$ is the sum  of the identity and two 1-sparse matrices, hence computing $R_{MZ}$ requires only $\mathcal{O}( (n+m) \cdot M)$ operations.

With our considerations from above we can write the optimal coefficients as
\begin{align}
    &\tilde{\alpha}^* = \left(R_{MZ} R_{MZ}^\top + (n+m)\lambda K_{MM}  \right)^{-1} K_{MZ}\delta,\\
    &\Leftrightarrow  \left(R_{MZ} R_{MZ}^\top + (n+m)\lambda K_{MM}  \right) \tilde{\alpha}^* = K_{MZ}\delta \label{eq:falkon_ready}
\end{align}
Computing $R_{MZ} R_{MZ}^\top$ explicitly costs $\mathcal{O}((n+m) M^2)$ operations and would thus dominate the cost of our previous operations. However, \eqref{eq:falkon_ready} is now exactly in the same form as Eq.~(8) in \citet{Rudi2017}. Thus from this point onwards we can build on their results to efficiently find a solution.

The key idea of \citet{Rudi2017} is to find an efficient way to precondition the system of linear equations in \eqref{eq:falkon_ready}. In analogy, we propose to use the following preconditioner
\begin{align}\label{eq:preconditioner}
    BB^\top = \left( \frac{n+m}{M} R_{MM}R_{MM}^T + \lambda (n+m) K_{MM} \right)^{-1},
\end{align}
where $R_{MM} := K_{MM}N_M$ and $N_M$ is defined in analogy to $N$ but only with the $M$ Nyström centers. The preconditioner \eqref{eq:preconditioner} thus corresponds to the ideal preconditioner of the problem without Nyström approximation but only $M$ points to start with.

Using this preconditioner we use $t$ conjugate gradient steps to solve
\begin{align}
    B^\top\left(R_{MZ} R_{MZ}^\top + (n+m)\lambda K_{MM}  \right) B\beta = B^\top K_{MZ}\delta.
\end{align}
If $\hat{\beta}$ is the approximate solution after $t$ steps, we obtain an approximate solution as
\begin{align}
    \hat{\alpha} = B \hat{\beta}.
\end{align}
The algorithm is described in Algorithm \ref{alg:falkon} and has overall complexity of $\mathcal{O}((\ntr+\mtr )Mt + M^3)$ in time and $\mathcal{O}(M^2)$.
\end{document}